\documentclass[a4paper,fleqn]{cas-dc}

\usepackage[numbers]{natbib}
\usepackage{amsmath,amsfonts}
\usepackage{algorithmic}
\usepackage{algorithm}

\usepackage{multirow}
\usepackage{subcaption}
\usepackage{bigstrut}
\usepackage{colortbl}
\usepackage{xcolor}
\usepackage{overpic}
\usepackage{booktabs}
\usepackage[capitalize]{cleveref}
\crefname{section}{Sec.}{Secs.}
\Crefname{section}{Section}{Sections}
\Crefname{table}{Table}{Tables}
\crefname{table}{TAB.}{TABS.}

\usepackage{soul}
\graphicspath{{figures/}}

\def\tsc#1{\csdef{#1}{\textsc{\lowercase{#1}}\xspace}}
\tsc{WGM}
\tsc{QE}
\begin{document}
\let\WriteBookmarks\relax
\def\floatpagepagefraction{1}
\def\textpagefraction{.001}

\shorttitle{Dense Affinity Matching for Few-Shot Segmentation}    

\shortauthors{Hao Chen et al.}  

\title [mode = title]{Dense Affinity Matching for Few-Shot Segmentation}  

% % Title footnote mark
% % eg: \tnotemark[1]
% \tnotemark[<tnote number>] 

% % Title footnote 1.
% % eg: \tnotetext[1]{Title footnote text}


\author[inst1]{Hao Chen}[orcid=0000-0002-5700-1202]
\author[inst2]{Yonghan Dong}
\author[inst1]{Zheming Lu}

\author[inst1]{Yunlong Yu}
\fnmark[*]
\ead{yuyunlong@zju.edu.cn}
\cortext[1]{Corresponding author}

\author[inst1]{Yingming Li}
\author[inst3]{Jungong Han}
\author[inst4]{Zhongfei Zhang}

% % Corresponding author indication
% \cormark[<corr mark no>]

% % Footnote of the first author
% \fnmark[<footnote mark no>]

% % Email id of the first author
% \ead{<email address>}

% % URL of the first author
% \ead[url]{<URL>}

% % Credit authorship
% % eg: \credit{Conceptualization of this study, Methodology, Software}
% \credit{<Credit authorship details>}

% \affiliation[<aff no>]{organization={},
% %          citysep={}, % Uncomment if no comma needed between city and postcode

\affiliation[inst1]{organization={Zhejiang University},%Department and Organization
            addressline={38, Zheda Road}, 
            city={Hangzhou},
            postcode={310027},
            country={P.R. China}}

\affiliation[inst2]{organization={Huawei Technologies Ltd},%Department and Organization 
            city={Shenzhen},
            country={P.R. China}}

\affiliation[inst3]{organization={Aberystwyth University},%Department and Organization
            addressline={E49 Llandinam Building}, 
            city={Aberystwyth},
            postcode={310027},
            country={UK}}

\affiliation[inst4]{organization={State University of New York},%Department and Organization
            addressline={Binghamton}, 
            city={NY},
            postcode={13902-6000},
            country={USA}}

% % Address/affiliation
% \affiliation[<aff no>]{organization={},
% %          citysep={}, % Uncomment if no comma needed between city and postcode

% \cortext[1]{Corresponding author}



\begin{abstract}
Few-Shot Segmentation (FSS) aims to segment the novel class images with a few annotated samples. In this paper, we propose a dense affinity matching (DAM) framework to exploit the support-query interaction by densely capturing both the pixel-to-pixel and pixel-to-patch relations in each support-query pair with the bidirectional 3D convolutions. Different from the existing methods that remove the support background, we design a hysteretic spatial filtering module (HSFM) to filter the background-related query features and retain the foreground-related query features with the assistance of the support background, which is beneficial for eliminating interference objects in the query background. We comprehensively evaluate our DAM on ten benchmarks under cross-category, cross-dataset, and cross-domain FSS tasks. Experimental results demonstrate that DAM performs very competitively under different settings with only 0.68M parameters, especially under cross-domain FSS tasks, showing its effectiveness and efficiency.
\end{abstract}


% % Research highlights
% \begin{highlights}
% \item 
% \item 
% \item 
% \end{highlights}

% Keywords

\maketitle

\section{Introduction}
\label{sec:intro}

With the rapid development of computer vision, semantic segmentation \cite{fcn, u-net}, as one of the most important vision fields, has made remarkable progress. The great success of semantic segmentation benefits from a large amount of human-annotated datasets. However, the pixel-level annotations are hard to obtain due to the time-consume and labor. To alleviate this problem, Few-Shot Segmentation (FSS) \cite{sagnn, catrans, MAO2022104, PUTHUMANAILLAM2023126201, WANG20231} aims at segmenting samples with a few annotated support samples and has been attracting a lot of attention.

FSS remains a very challenging task due to the scarcity of support samples and the diverse intra-class flavors. The crux of FSS is to exploit the affinities among the objects in the support-query pairs. Currently, the existing approaches roughly follow two groups, i.e., class-wise and pixel-wise, based on the representations of support samples. The class-wise methods \cite{asgnet, dpcn, cwt, fptrans_zhang2022featureproxy, jiao2022mask} perform the support masks on the feature maps of the support samples to obtain their foreground prototype vectors and utilize them to guide the segmentation of the query images. As the compressed prototype vectors only contain the most manifest information while losing the spatial structure information that is essential for the dense FSS task, the class-wise methods fail in conducting fine-grained matches with target objects in the query image. To remedy the spatial information loss, the pixel-wise methods \cite{hsnet,dcama} represent the support-query pairs with pixel-wise features and obtain the support-query affinity by performing the dense many-to-many correspondence.

\begin{figure}[t]
  \centering
   \includegraphics[width=0.98\linewidth]{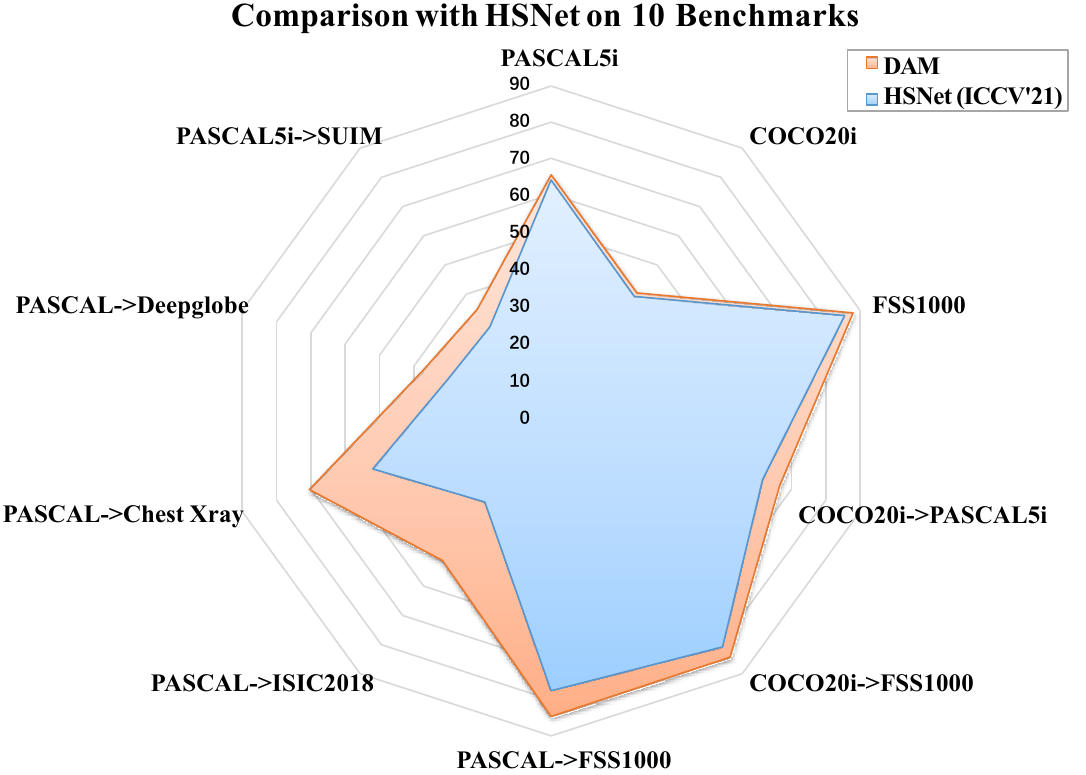}
   \caption{Comprehensive comparison results(\%) between HSNet \cite{hsnet} and DAM on 10 benchmarks, where each node indicates mIoU in the different benchmark. All experiments are tested with Resnet50 backbone under the 1-Shot setting. The experiment results of HSNet are collected from \cite{hsnet,rtd_Wang_2022_CVPR,patnet}. }
   \label{fig:radar}
   \vspace{-0.3cm}
\end{figure}

% segmenting the query samples from the naive correlations directly obtained with the support-query feature maps

Though pixel-wise methods achieve superior performances, segmenting the query samples from the support-query correlation matrix once obtaining it would lead to inferior relation matching due to the following reasons. First, the pixel-to-pixel correlation hardly handles the case where the objects in support-query pairs are in different sizes. Second, the foreground objects and background objects may give high relevance but the parts of the foreground objects may get low relevance due to the inter-class similarity and intra-class diversity, which will mislead the model in incomplete query objects and discovering interference objects in the background. Third, most existing approaches remove the support backgrounds with the support mask in advance and only consider the object foregrounds in the dense correlations, which will omit some important information for the query segmentation.

To address the above issues, we develop a dense affinity matching (DAM) framework for FSS to enhance the correspondence matching from the naive support-query affinity by fully exploiting the dense pixel-to-patch relations with feasible sizes between the foreground and background. Different from HSNet \cite{hsnet} which removes the support background with the support mask before extracting the support-query correlations, DAM obtains the support-query affinity from the whole feature maps of support-query pairs, which fully considers the support background. Denser support-query affinity is, more foreground-related features will be retained and more background-related features will be filtered in the follow-up hysteretic spatial filtering module, thus leading to a better prediction of query mask. In contrast to DCAMA \cite{dcama} that weights the support-query correlation with the support mask once obtaining it, we propose a hysteretic spatial filtering module (HSFM) to further enhance the support-query correlation with bidirectional 3D convolutions before filtering it. Specifically, the bidirectional 3D convolutions exploit the support-query correlations in both pixel-to-pixel and pixel-to-patch with flexible size, reaching more dense correlations and more fine-grained matching between the support-query pairs, which contributes to segmenting the query target objects with the huge different size from the support objects. After the support-query affinity enhancement, HSFM filters the query background-related features and retains the foreground-related features with the assistance of the support mask.

 Besides, the existing dense pixel-level approaches (e.g., DCAMA \cite{dcama}) add a linear head on the top of each block of the backbone to reduce the noises before obtaining the affinity matrix. Though effective it is, the linear head introduces a large number of parameters, resulting in a heavy segmentation head. To lighten the model, we remove the linear head and mitigate the noises after obtaining the affinity matrix. With more convolutional layers produced in the affinity matrix, DAM fuses multi-scale and multi-receptive-field feature maps with fewer parameters, which provides the learnable space for reducing the noises.

The main contributions of this work include:

\begin{itemize}
    \item We propose a lightweight FSS framework, Dense Affinity Matching (DAM), with  0.68M parameters, to boost the support-query correspondence matching by fully exploiting the dense affinity matching of both the background and foreground in each support-query pair. 

    \item We develop a hysteretic spatial filtering module (HSFM) to further enhance the support-query affinity with bidirectional 3D convolutions and filter it with the support mask, which exploits both the pixel-to-pixel and pixel-to-patch relationships.

    \item As far as we know, this is the first work to comprehensively evaluate FSS under cross-category, cross-dataset, and cross-domain settings. The experiments on ten benchmarks show the superior generalization ability of our method. As shown in \cref{fig:radar}, DAM comprehensively outperforms HSNet \cite{hsnet} under different settings.
    
\end{itemize}

\section{Related work}\label{sec2}

\begin{figure*}[t]
  \centering
   \includegraphics[width=0.9\linewidth]{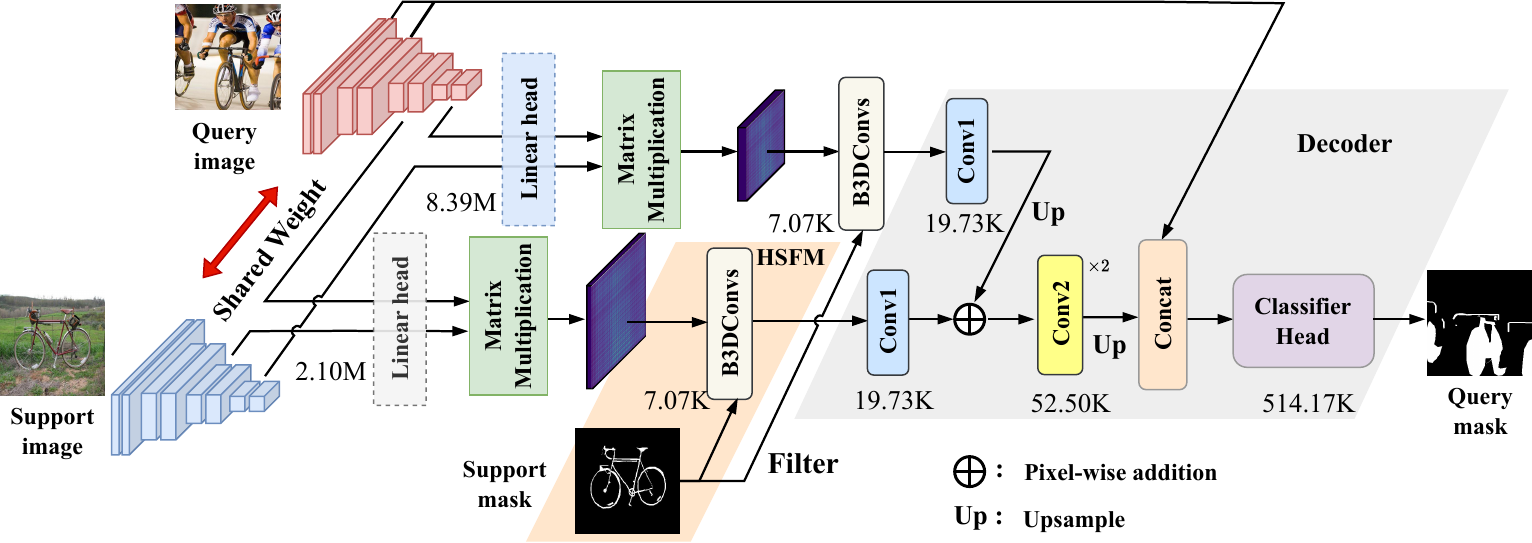}
   \caption{Illustration of the proposed dense affinity matching(DAM) framework. DAM extracts the multi-scale features from the frozen backbone and reaches the support-query affinity matrix between query features and support features from each layer of the last two backbone blocks. Then, the Hysteretic Spatial Filtered Module (HSFM) filters the query background and retains the query foreground with the support mask. In the HSFM, the query-support affinity matrix is further enhanced with the proposed bidirectional 3D convolutions (B3DConvs). Finally, the filtered query-support affinity matrix from different blocks combined with the low-level query features are used to predict the query mask with a decoder. The `\textbf{dotted line}' denotes that the linear head is removed in the framework. The number next to each module denotes its parameter.} 
   \label{fig:framework}
   \vspace{-0.2cm}
\end{figure*}

\subsection{Few-Shot Learning}
As deep learning advances, researchers strive to maximize the utility of each example when dealing with limited labeled data. One approach to tackle this issue is Few-Shot Learning (FSL) \cite{XIE2023126426_fsl, ZHAO2022412_vr, QIN202269_kdm}, which aims to learn novel concepts using a minimal number of annotated samples. Earlier FSL research primarily focused on classification tasks \cite{ZHANG2022406_fsc}, but recent developments have significantly expanded its applications to include object detection \cite{DU202213_fsod, ZHANG2022216_fsod} and semantic segmentation\cite{imtfa, svf, dgpnet, PUTHUMANAILLAM2023126201, WANG20231}.

\subsection{Fow-Shot Semantic Segmentation}

The existing FSS works \cite{mmnet_Wu_2021_ICCV,  rpnet_Tang_2021_ICCV, moon2022msi, min2023hybrid} could roughly follow two groups, i.e., class-wise \cite{oslsm,sgone,miningfss_Yang_2021_ICCV, tian2022capl} and pixel-wise \cite{asnet, cats} correspondence, based on the interaction between the support-query pairs. The class-wise approaches represent the objects in the support samples as a prototype through masked average pooling and segment the query samples based on the similarities between the support prototype and the pixel feature embeddings of the query sample. The class-wise approaches differ in the way of obtaining the support prototypes. For example, ASGNet \cite{asgnet} extracts multiple prototypes via clustering and adaptively allocates these prototypes to the most related query pixels, DPCN \cite{dpcn} introduces a dynamical convolutional module to extract prototypes containing support object details and extracts query foreground through feature fusion, and NTRENet \cite{ntrenet} extracts the background prototype to eliminate the similar query region in a prototype-to-pixel way. Though efficient, the class-wise approaches compress the support feature maps into a prototype vector and lose their spatial information which is essential for the segmentation tasks. 

To remedy the spatial information loss, the pixel-wise methods obtain the support-query correlation by densely calculating the similarities between the pixel feature embeddings of support-query pairs. Though more operations than the class-wise competitors, the pixel-wise approaches obtain superior performances and gain much more attention in recent years. These approaches boost the performances by enhancing the support-query correlations. For example, as one of the earliest works to compute pixel-to-pixel correlation, DAN \cite{dan} enhances the support-query correlation with a graph attention mechanism. HSNet \cite{hsnet} enhances support-query correspondence with 4D convolution operations. Following HSNet \cite{hsnet}, VAT \cite{vat} replaces the 4D convolution operations with a 4D swin transformer \cite{liu2021swin} to further enhance the support-query correspondence. DCAMA \cite{dcama} enhances the support-query correlations by fully aggregating the pixel-to-pixel relations of all layers of each backbone block. Our DAM is related to HSNet \cite{hsnet} and DCAMA \cite{dcama} and enhances the support-query correlation by first densely calculating the pixel-to-pixel relations and then further processing it with bidirectional 3D convolutions. Different from HSNet and DCAMA that filter the support background\cite{hsnet} or weight the affinity matrix with the support mask once obtaining it\cite{dcama}, DAM introduces a hysteretic spatial filtering module to fully consider the pixel-to-pixel and pixel-to-patch relationships of background and foreground between the support-query pairs before filtering it. 

Many works have explored the effects of support backgrounds. BAM \cite{bam} focuses on the interference brought by the support background and designs a branch to learn the base category distribution. HM \cite{hm} compares the difference between the backbone features with or without the background of the input image and complements each other. Different from these methods, DAM focuses on the affinity matching of background in a more fine-grained way, which contributes to filtering similar objects in query samples.

\subsection{Cross-Domain Few-Shot Segmentation}

Cross-Domain Few-Shot Segmentation is a specific FSS scenario that performs the trained model on the novel classes from a novel domain. In contrast to ordinary FSS, the cross-domain FSS is more challenging due to the shortage of the prior distribution of the testing set. There are a few attempts to address cross-domain FSS. RTD \cite{rtd_Wang_2022_CVPR} designs a two-stage method to transfer the feature-enhancement knowledge to target samples. PATNet \cite{patnet} measures the cross-domain tasks difficulty and proposes a pyramid-based module to transform the domain-specific features into domain-agnostic ones. In this work, we extensively evaluate DAM in cross-domain tasks and show that DAM has a great transfer ability between different domains.

\section{Methods}\label{sec3}

\subsection{Problem Definition}

Given a base set consisting of some images and their masks, FSS aims to train a model to segment some samples from the novel classes, under the condition of one or a few support samples associated with their masks. In this work, we adopt the popular meta-learning paradigm to train the model with the episodes sampled from the base set, where each episode contains a support set $\mathcal{S}={\{ I_{k}^{s},\ M_{k}^{s}\}}_{k = 1}^{K}$ and a query set $\mathcal{Q} = \{ I^{q}, M^{q} \}$, $I^s$ and $I^q$ represents the input support and query images, $M^s$ and $M^q$ denotes the corresponding masks, and $K$ denotes the number of support images. $M^{q}$ is required to be predicted during inference.

\subsection{Framework}

The FSS task is to match the objects from the same category in the support-query pairs. In this paper, we propose a novel framework to address FSS by densely exploiting pixel-level support-query pairs. As illustrated in \cref{fig:framework}, the framework could be divided into four steps. (1) The support-query pair is taken as the input to a feature encoder (e.g., ResNet50) to obtain their feature maps from different blocks. (2) The feature maps from the last two blocks are used to reach the support-query affinities. Same as the pixel-wise methods \cite{hsnet, dcama}, we compute the support-query correlations with the fully pixel-to-pixel inner product of feature maps from each layer in blocks and concatenate them separately. (3) The affinity matrix is filtered through the hysteretic spatial filtering module (HSFM) with the corresponding support mask, which contributes to obtaining the coarse mask predictions of the query sample by filtering the query background and retaining the query foreground. In the HSFM, the support-query affinities are enhanced with the bidirectional 3D convolutions that could further exploit the pixel-to-patch relationships, contributing to segmenting the query objects with different sizes from the support objects. (4) The coarse mask predictions are refined in a decoder network to predict the query mask with the low-level query feature embeddings from the first two backbone blocks.

Steps (3) and (4) are key designs in our framework, which will be introduced in detail. 

\begin{figure}[t]
  \centering
   \includegraphics[width=0.98\linewidth]{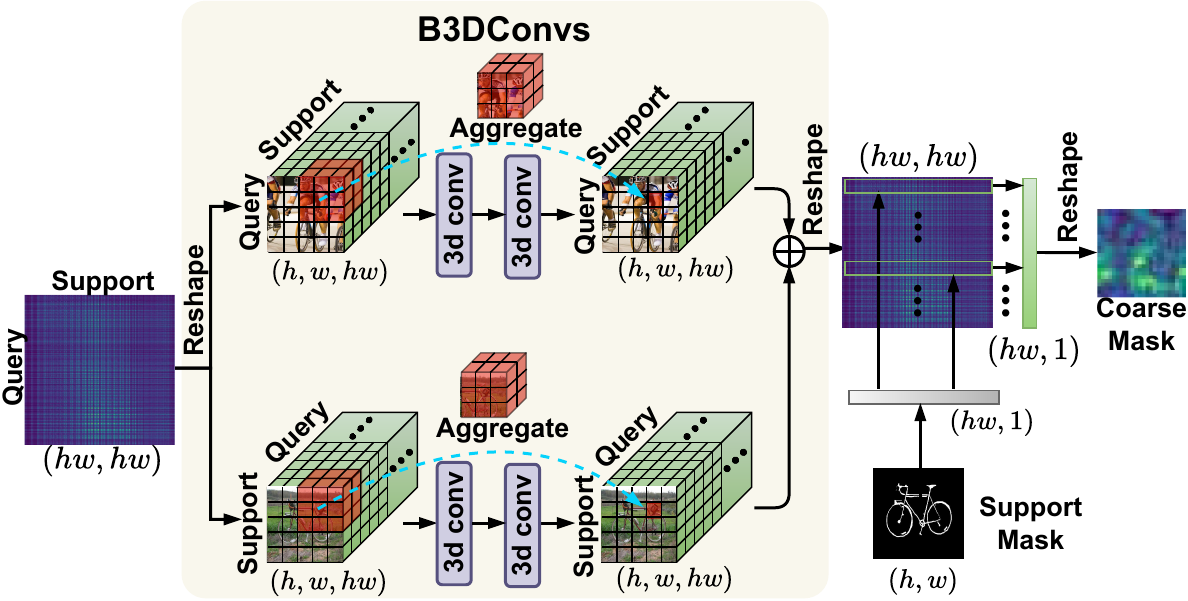}
   \caption{Illustration of the proposed Hysteretic Spatial Filtering Module (HSFM). Bidirectional 3D Convolutions (B3DConvs) aim to enhance the support-query affinity. Then, the enhanced support-query affinity is filtered by suppressing the query background and retaining the query foreground with the assistance of the support mask. }
   \label{fig:filter}
\end{figure}

   % The reshaped correlation matrix is enhanced in support and query parts separately and fused by pixel-wise addition. Then, the coarse mask is filtered by the reshaped support mask.

\subsection{Hysteretic Spatial Filtering Module}

Given the feature maps $F_s \in \mathbb{R}^{c\times h \times w}$ and $F_q \in \mathbb{R}^{c\times h \times w}$ of both support and query samples, where $c$, $h$, and $w$ denote the channel, height, and width, respectively, the support-query affinity matrix $S\in \mathbb{R}^{hw\times hw}$ could be obtained with 
\begin{equation} \label{affinity}
    S = R(F_s)^T \times R(F_q),
\end{equation}
where $R$ denotes the reshaping operation that reshapes the feature maps of support and query samples to $\mathbb{R}^{c\times hw}$. To further enhance the affinity matrix, we introduce the bidirectional 3D convolutions (B3DConvs) on it to exploit the pixel-to-patch correlations between support-query pairs. As shown in \cref{fig:filter}, we first reshape $S$ to $\mathbb{R}^{h \times w\times hw}$ and input it to a 3D convolution network with different kernels. Each network contains two 3D convolution blocks, consisting of a 3D convolutional layer, a BatchNorm layer, and a ReLU operation. Thus, each pixel of the support sample could interact with the query patches whose size is the same as the kernels (e.g., $3\times 3$ and $5\times 5$ ). Similarly, we could explore the interactions of each query pixel and the support patches by transposing and reshaping $S$ and then inputting it into the 3D convolution network. In this way, we could obtain the pixel-to-patch relationships in a \textit{bidirectional} way, i.e., support-to-query, and query-to-support. This process is formulated as:
\begin{equation} \label{b3d}
    S_{enh} = R_2(H(R_1(S)) ) + R_2( H(R_1(S^T)) )^T,
\end{equation}
where $S_{enh} \in \mathbb{R}^{hw \times hw}$ denotes the enhanced affinity matrix, $H$ denotes a 3D convolution network, $R_1$ and $R_2$ are both shaping operations, $T$ is the metric transpose.   

Once obtained the enhanced support-query affinity matrix, we introduce the support mask to filter the query foreground. As shown in \cref{fig:filter}, the reshaped support mask $M_s \in \mathbb{R}^{hw \times 1} $  performs as a convolution operator on the affinity matrix along the query dimension to obtain the coarse query mask, which is formulated as:
\begin{equation}
    % M_{coarse} = S_{enh} \odot M_s,
    M_{coarse} = Conv(input: S_{enh}, weight: M_s)
\end{equation}
where $M_{coarse}\in \mathbb{R}^{hw \times 1}$ denotes the coarse mask of query sample, $Conv$ denotes the convolution operation.

\begin{figure}[t]
  \centering
  \setlength{\abovecaptionskip}{0.4cm}
  \begin{overpic}[width=1.0\linewidth]{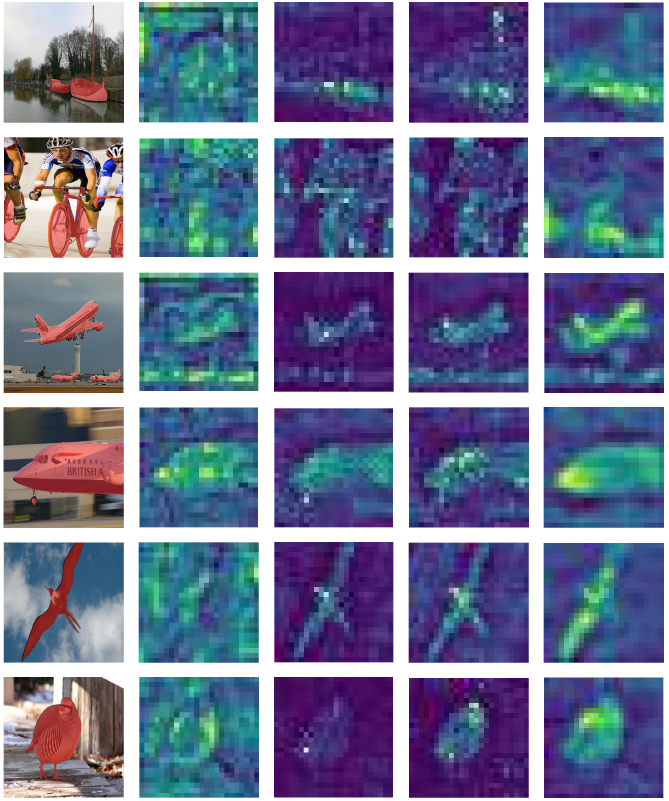}
  \small
  \put(4,-2){Query}
   \put(21,-2){HSNet}
   \put(35,-2){DCAMA$^\ast$}
   \put(52,-2){DCAMA}
   \put(67,-2){DAM(Ours)}
  \end{overpic}
   \caption{The visualization of the coarse mask in the second block of different models, trained under 1-shot task with Resnet50 backbone on PASCAL$5^i$ dataset. $\ast$ denotes DCAMA with the last two backbone blocks. }
   \label{fig:feature-map}
\end{figure}

In contrast to DCAMA \cite{dcama} which weights affinity matrix with the support mask once obtaining it, DAM introduces a hysteretic spatial filtering module, which takes the affinity matrix as two parts and produces 3D convolutions to them separately before filtering. Through a fine-grained operation, our method guarantees the affinity matching of the foreground and background, benefiting suppressing the related background and highlighting the related foreground in spatial filtering. \cref{fig:feature-map} provides some filtered feature maps of different approaches and shows that DAM retains more query foreground and filters more query background, which will benefit in the following mask prediction.

\begin{figure}[t]
	\centering
	\subfloat[Conv1]{\includegraphics[width=3in]{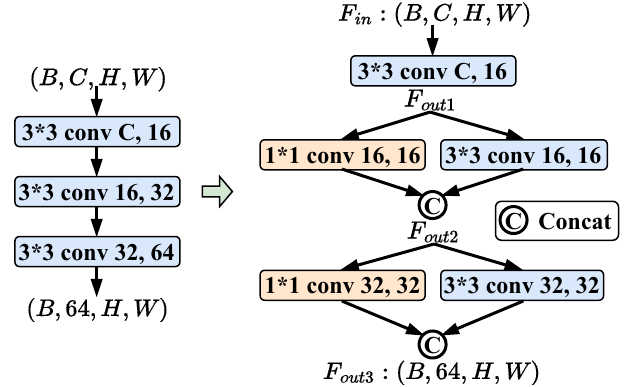}%
		\label{conv1}}
	\hfil
	\vspace{2mm}
	\subfloat[Conv2]{\includegraphics[width=3in]{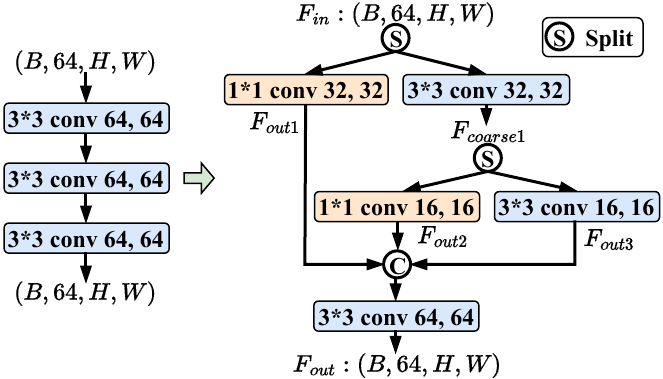}%
		\label{conv2}}
	\caption{Illustration of the proposed Conv1 and Conv2 modules in the decoder. The right convolutional networks are strengthened from the left ones, which contain more convolutional layers and fewer parameters.}
	\label{fig:conv}
\end{figure}

\subsection{Lightweight Designs}

To fuse the multi-scale and multi-receptive-field coarse masks, DAM redesigns the convolutional operations to increase the convolutional layers with fewer parameters. During the framework, the coarse masks of two blocks are enhanced with Conv1 separately and fused by pixel-wise addition. The fused feature is continuously enhanced with two Conv2 blocks.

Conv1 and Conv2 are two convolutional blocks in the decoder module. As shown in \cref{fig:conv}, two multi-scale and multi-receptive-field convolutional networks \textit{(in \cref{fig:conv} (a), (b) right)} are introduced to replace the ordinary convolutional networks \textit{(in \cref{fig:conv} (a), (b) left)}. Each \textit{conv block} contains a 2d convolution layer, a GroupNorm layer, and a ReLU layer. For \textit{Conv1}, the feature map is enhanced with a $1\times1$ \textit{conv block} and a $3\times3$ \textit{conv block}. And the outputs of two \textit{conv blocks} are concatenated in the channel dimension for multi-receptive-field features fusion, which is formulated as:
\begin{equation}\label{equ:conv1}
\begin{split}
   F_{out1} &= C_{3\times3}(F_{in}), \\
   F_{out2} &= Concat(C_{1\times1}(F_{out1}), C_{3\times3}(F_{out1})), \\
   F_{out3} &= Concat(C_{1\times1}(F_{out2}), C_{3\times3}(F_{out2})),
\end{split}
\end{equation}
where $Concat$ denotes concatenating two feature maps in the channel dimension, $C_{1\times1}$ and $C_{3\times3}$ denote a convlutional layer with the $1 \times 1$ and $3 \times 3$ kernels, separately.

For \textit{Conv2}, the feature map $F_{in}$ is split into two parts in the channel dimension, which is processed by a $1\times1$ \textit{conv block} and a $3\times3$ \textit{conv block} respectively, obtaining $F_{out1}$ and $F_{coarse1}$. And the feature map $F_{coarse1}$ is split and processed by the mentioned two \textit{conv blocks} again, obtaining $F_{out2}$ and $F_{out3}$. Then, all the output features are concatenated in the channel dimension to match the size of the input feature $F_{in}$. The process is formulated as:

\begin{equation}\label{equ:conv2}
\begin{split}
   &F_{out1}, F_{coarse1} = Split(F_{in}), \\
   &F_{out1}, F_{coarse1} = C_{1\times1}(F_{out1}), C_{3\times3}(F_{coarse1}), \\
   &F_{out2}, F_{out3} = Split(F_{coarse1}), \\
   &F_{out2}, F_{out3} = C_{1\times1}(F_{out2}), C_{3\times3}(F_{out3}), \\
   &F_{out} = Concat(F_{out1}, F_{out2}, F_{out3}),
\end{split}
\end{equation}
where $Split$ denotes splitting the feature map in the channel dimension, $Concat$ denotes concatenating feature maps in the channel dimension, $C_{1\times1}$ and $C_{3\times3}$ denote a convlutional layer with the $1 \times 1$ and $3 \times 3$ kernels, separately.

As shown in \cref{fig:conv}, the \textit{right} convolutional network, with fewer parameters, consists of more convolutional layers than the \textit{left} one. With $Conv1$ and $Conv2$, the multi-scale and multi-receptive-field features are fused, which is important for the pixel-level segmentation task. Moreover, the increased convolutional layers provide the learning space for the affinity matrix to eliminate the noises brought by the freeze backbone, which benefits segmenting the query samples.

In \cite{dcama,cyctr}, a linear head is added for each of the last two modules to transform the feature embeddings into the other feature embedding to better fit the segmentation task from the pre-trained backbone. Though the linear head could assist the model in improving the cross-category FSS performance, it consists of a large number of learnable parameters. To lighten the model, we remove the linear heads as shown in \cref{fig:framework} and remedy it with the post-processing in the decoder with fewer parameters. Besides, we observe that removing the linear heads could preserve more domain-invariant features, which will benefit the cross-domain FSS tasks.

\begin{figure}[t]
  \centering
  \includegraphics[width=0.8\linewidth]{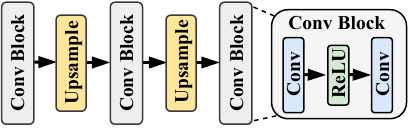}
  \caption{Illustration of the Classifier Head in the decoder.}
  \label{fig:classifier-head}
\end{figure}

\subsection{Classifier Head}

The classifier head is designed to enhance the query foreground information in the decoder module. As shown in \cref{fig:classifier-head}, Classifier Head contains a series of convolutional blocks and two upsampling operations between two convolutional blocks. Each convolutional block consists of two convolutional layers connected with a ReLU operation.

% \subsection{Decoder}
% The decoder takes the coarse query masks from different layers as the input to predict the final mask. Specifically, the decoder fuses the multi-scale and multi-receptive-field coarse masks with Conv1 and Conv2. The coarse masks of two blocks are enhanced with Conv1 separately and fused by pixel-wise addition. The fused feature is continuously enhanced with two Conv2 blocks. To supplement the structural information, the feature maps are concatenated with the low-level query features and segmented with the classifier head. 

% In \cite{dcama,cyctr}, a linear head is added for each of the last two modules to transform the feature embeddings into the other feature embedding to better fit the segmentation task from the pre-trained backbone. Though the linear head could assist the model in improving the cross-category FSS performance, it consists of a large number of learnable parameters. To lighten the model, we remove the linear heads as shown in \ref{fig:framework} and remedy it with the post-processing in the decoder with fewer parameters. Besides, we observe that removing the linear heads could preserve more domain-invariant features, which will benefit the cross-domain FSS tasks, please refer to more details in Supplementary Material. 
During training, the binary cross-entropy loss is used to supervise the model. Under the 5-shot setting, multiple support features and masks are concatenated in the channel dimension in Hysteretic Spatial Filtering Module. 

\begin{figure}[t]
  \centering
  \setlength{\abovecaptionskip}{0.4cm}
  \begin{overpic}[width=1.0\linewidth]{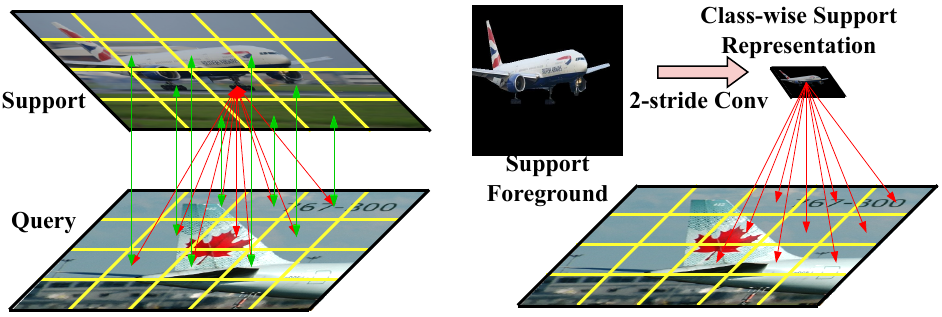}
  \small
   \put(16,-3){(a) DAM}
   \put(68,-3){(b) HSNet}
  \end{overpic}
  \vspace{-1mm}
   \caption{The structure comparison of DAM and HSNet. }
   \label{fig:damvshsnet}
\end{figure}

\subsection{Compared with HSNet}

In the paper, we set HSNet \cite{hsnet} as the major competitor to our DAM. DAM achieves better performance in the 10 mentioned benchmarks compared to HSNet, with fewer parameters, 0.68M, especially in cross-domain tasks. The main difference is located in the following four aspects:

\textbf{Structure.} As shown in \cref{fig:damvshsnet}, HSNet extracts a class-wise feature representation from the foreground of the support images with a 2-stride convolutional operation and performs query image segmentation by matching the extracted class-wise support representation and the query features. In contrast, DAM focuses on extracting the pixel-to-pixel and pixel-to-patch correlations between the foreground and background of support-query pairs.

\textbf{Method.} HSNet extracts class-wise support representations, which are intrinsically class-related and unlikely to transition across different domains. DAM prioritizes pixel-wise correlations that are class-agnostic due to the non-categorical nature of single pixels, which benefits tackling more challenging cross-domain tasks.

\textbf{Support Background Exploiting.} Regarding the handling of support backgrounds, HSNet removes them before extracting affinity matching, while our DAM incorporates a learnable HSFM that aggregates support-query affinity with the support masks. This technique sets DAM apart from other methods and contributes to its unique approach to general tasks.

\textbf{Efficiency and Accuracy.} DAM achieves better performance in 10 benchmarks with fewer parameters than HSNet, especially in cross-domain tasks.

% As shown in \cref{fig:damvshsnet}, HSNet extracts a class-wise feature representation from the foreground of the support images with a 2-stride convolutional operation and performs query image segmentation by matching the extracted class-wise support representation and the query features. In contrast, DAM focuses on extracting the pixel-to-pixel and pixel-to-patch correlations between the foreground and background of support-query pairs. Regarding the handling of support backgrounds, HSNet removes them before extracting affinity matching, while our DAM incorporates a learnable HSFM that aggregates support-query affinity with the support masks. 

% HSNet extracts class-wise support representations, which are intrinsically class-related and unlikely to transition across different domains. DAM prioritizes pixel-wise correlations that are class-agnostic due to the non-categorical nature of single pixels, which benefits tackling more challenging cross-domain tasks.

\section{Experiments}
\label{sec:experiments}

\begin{table*}[t]
  \centering
  \caption{FSS performances (\%) on PASCAL-5$^i$ and COCO-20$^i$ with different backbones (ResNet50 and ResNet101). `$\ast$' denotes the results obtained by ourselves with the released codes. The results of all the competitors are from the published literature. The best and the second best results are marked in \textbf{bold} and \underline{underline}, respectively.}
  \makebox[\textwidth]{
  \resizebox{\linewidth}{!}{
    \begin{tabular}{p{1em}|l|c|cccc|c|cccc|c|c}
    \toprule
    % \hline
    \multicolumn{14}{c}{Pascal-5$^i$} \bigstrut\\
    \midrule
    % \hline
    \multicolumn{1}{p{1em}|}{\multirow{2}[4]{*}{Backbone}} & \multirow{2}[4]{*}{Methods} & \multicolumn{1}{c|}{\multirow{2}[4]{*}{Type}} & \multicolumn{5}{c|}{1-shot}           & \multicolumn{5}{c|}{5-shot}           & \multicolumn{1}{p{3.875em}}{Learnable} \bigstrut\\
\cline{4-13}          & \multicolumn{1}{c|}{} &       & \multicolumn{1}{p{2.75em}}{Fold-0} & \multicolumn{1}{p{2.75em}}{Fold-1} & \multicolumn{1}{p{2.75em}}{Fold-2} & \multicolumn{1}{p{2.75em}|}{Fold-3} & \multicolumn{1}{p{3em}|}{mIoU} & \multicolumn{1}{p{2.75em}}{Fold-0} & \multicolumn{1}{p{2.75em}}{Fold-1} & \multicolumn{1}{p{2.75em}}{Fold-2} & \multicolumn{1}{p{2.75em}|}{Fold-3} & \multicolumn{1}{p{3em}|}{mIoU} & \multicolumn{1}{p{3.875em}}{Params(M)} \bigstrut\\
    % \hline
    \midrule
          % & PPNet(ECCV'20)\cite{ppnet} &       & 48.6  & 60.6  & 55.7  & 46.5  & 52.8  & - & 58.9  & 68.3  & 66.8  & 57.98 & 63.0   & - &  31.5 \\
          % & RPMM(ECCV'20)\cite{pmm} &       & 55.2  & 66.9  & 52.6  & 50.7  & 56.3  & - & 56.3  & 67.3  & 54.5  & 51.0  & 57.3  & - & - \\
          % & RePRI(CVPR'21)\cite{repri} &       & 60.2  & 67.0  & 61.7  & 47.5  & 59.1  & 64.5  & 70.8  & \underline{71.7} & 60.3  & 66.8  & - \\
          & CWT(ICCV'21)\cite{cwt} &       & 56.3  & 62.0  & 59.9  & 47.2  & 56.4  & 61.3  & 68.5  & 68.5  & 56.6  & 63.7  & 47.3 \\
          % & CAPL(CVPR'22) & \multicolumn{1}{p{3.71em}|}{Prototype} & - & - & - & - & 62.2  & - & - & - & - & - & 67.1  & - & - \\
          & DCP(IJCAI'22)\cite{dcp} &  & 63.8 & 70.5 & 61.2 & 55.7 & 62.8  & 67.2 & 73.2 & 66.4 & 64.5 & 67.8 & 11.3 \\
          & NTRENet(CVPR'22)\cite{ntrenet} & \multicolumn{1}{p{3.71em}|}{Prototype} & 65.4  & 72.3  & 59.4  & 59.8  & 64.2   & 66.2  & 72.8  & 61.7  & 62.2  & 65.7  & 18.6 \\
          & SSP(ECCV'22)\cite{ssp} &       & 60.5  & 67.8  & \textbf{66.4} & 51.0 & 61.4  & 67.5  & 72.3  & \textbf{75.2}  & 62.1  & 69.3  & 8.7 \\
          & IPMT(NeurIPS'22)\cite{ipmt_liu2022intermediate} &       & \textbf{72.8}  & \textbf{73.7}  & 59.2 & \textbf{61.6}  & \textbf{66.8}  & \underline{73.1}  & \textbf{74.7}  & 61.6  & 63.4  & 68.2  & - \\
          % & RPMG-FSS\cite{rpmg-fss}(TCSVT'23) &       & 64.4  & 72.6 & 57.9  & 58.4  & 63.3  & 65.3  & 72.8    & 58.4  & 59.8  & 64.1  & 36.98 \\
          & TAAM(Neurocom'22)\cite{MAO2022104} &       & 59.4  & 70.5 & 61.0  & 57.1  & 62.0  & 64.3  & 71.9    & 65.8  & 59.1  & 65.3  & - \\
\cmidrule(r){2-14} 
\multicolumn{1}{p{5em}|}{ResNet-50} & CyCTR(NeurIPS'21)\cite{cyctr} &       & 65.7  & 71.0 & 59.5  & 59.7  & 64.0  & 69.3  & 73.5  & 63.8  & 63.5  & 67.5  & 15.9$^\ast$ \\
          & HSNet(ICCV'21)\cite{hsnet} &       & 64.3  & 70.7  & 60.3  & \underline{60.5}  & 64.0  & 70.3  & 73.2  & 67.4  & \underline{67.1}  & 69.5  & 2.6 \\
          & AAFormer(ECCV'22)\cite{aaformer} &       & \underline{69.1}  & 73.3 & 59.1  & 59.2  & 65.2  & 72.5  & \textbf{74.7} & 62.0  & 61.3  & 67.6  & - \\
          & VAT(ECCV'22)\cite{vat} & \multicolumn{1}{p{3.71em}|}{Pixelwise} & 67.6  & 72.0  & 62.3  & 60.1  & 65.5  & 72.4  & 73.6  & 68.6  & 65.7  & \underline{70.1}  & 3.2 \\
          & DCAMA(ECCV'22)\cite{dcama} &  & 67.5  & 72.3  & 59.6  & 59.0  & 64.6  & 70.5  & 73.9  & 63.7  & 65.8  & 68.5 & 14.2$^\ast$ \\
          % & CMNet(TMM'22) &       & 65.4  & 71.5  & 55.2  & 58.1  & 62.5  & 73.5  & 67    & 71.7  & 55.8  & 59.9  & 63.6  & 74.1  &  \\
          % & QCLNet\cite{qclnet}(TCSVT'23) &       & 65.2  & 70.3  & 60.8  & \underline{61.0}  & 64.3  & 70.6  & 73.5    & 66.7  & \underline{67.1}  & 69.5  & 2.6 \\
          & ABCNet\cite{abcnet_Huang_2023_CVPR}(CVPR'23) &       & 68.8  & \underline{73.4}  & 62.3  & 59.5  & \underline{66.0}  & 71.7  & 74.2   & 65.4  & 67.0  & 69.6  & - \\
          & DAM (Ours) &       & 67.3  & 72.0  & \underline{62.4}  & 59.9  & 65.4  & \textbf{73.6} & \underline{74.6}  & \underline{69.9}  & \textbf{67.2}  & \textbf{71.3} & \textbf{0.68} \\
          % & \rowcolor{lightgray} Our\_standard &       & \textbf{70.3} & 72.1  & 62    & \textbf{62.8} & \textbf{66.8} & \textbf{78.5} & 72.7  & 73.2  & 68.8  & \textbf{69.5} & 71.1  & \textbf{81.9} & 42 \bigstrut[b]\\
\midrule

          % & PPNet(ECCV'20)\cite{ppnet} &       & 52.7  & 62.8  & 57.4  & 47.7  & 55.2  & - & 60.3  & 70.0 & 69.4  & 60.7  & 65.1  & -     &  50.5 \bigstrut[t]\\
          % & RePRI(CVPR'21)\cite{repri} &       & 59.6  & 68.6  & 62.2  & 47.2  & 59.4  & 66.2  & 71.4  & 67.0  & 57.7  & 65.6 & - \\
          & CWT(ICCV'21)\cite{cwt} &       & 56.9  & 65.2  & 61.2  & 48.8  & 58.0 & 62.6  & 70.2  & 68.8  & 57.2  & 64.7 & 66.3 \\
          % & CAPL(CVPR'22) & \multicolumn{1}{p{3.71em}|}{Prototype} & - & - & - & - & 63.6  & - & - & - & - & - & 68.9  & - & - \\
          & NTRENet(CVPR'22)\cite{ntrenet} & \multicolumn{1}{p{3.71em}|}{Prototype} & 65.5  & 71.8  & 59.1  & 58.3  & 63.7  & 67.9  & 73.2  & 60.1  & 66.8  & 67.0 & 18.6 \\
          & SSP(ECCV'22)\cite{ssp} &       & 63.2  & 70.4  & \textbf{68.5} & 56.3  & 64.6  & 70.5  & \underline{76.4} & \textbf{79.0} & 66.4  & \underline{73.1}  & 27.7 \bigstrut[b]\\
          & IPMT(NeurIPS'22)\cite{ipmt_liu2022intermediate} &       & \textbf{71.6} & \underline{73.5} & 58.0 & 61.2  & 66.1  & \textbf{75.3}  & \textbf{76.9} & 59.6 & 65.1  & 69.2  & - \\
          % & RPMG-FSS\cite{rpmg-fss}(TCSVT'23) &       & 63.0  & 73.3 & 56.8  & 57.2  & 62.6  & 67.1  & 73.3   & 59.8  & 62.7  & 65.7  & - \\
          
\cmidrule(r){2-14} 
\multicolumn{1}{p{5em}|}{ResNet-101}   & CyCTR(NeurIPS'21)\cite{cyctr} &       & 67.2  & 71.7  & 57.6  & 59.0  & 63.7 & 71.0  & 75.0  & 58.5  & 65.0  & 67.4  & 15.9$^\ast$ \bigstrut[t]\\
          & HSNet(ICCV'21)\cite{hsnet} &       & 67.3  & 72.3  & 62.0  & \underline{63.1} & 66.2  & 71.8  & 74.4  & 67.0 & 68.3  & 70.4  & 2.6 \\
          & AAFormer(ECCV'22)\cite{aaformer} &       & 69.9  & \textbf{73.6}  & 57.9  & 59.7  & 65.3  & \underline{75.0} & 75.1  & 59.0 & 63.2  & 68.1  & - \\
          & VAT(ECCV'22)\cite{vat} & \multicolumn{1}{p{3.71em}|}{Pixelwise} & 70.0 & 72.5  & 64.8  & \textbf{64.2}  & \underline{67.9}  & \underline{75.0} & 75.2  & 68.4  & \underline{69.5}  & 72.0  & 3.3 \\
          & DCAMA(ECCV'22)\cite{dcama} &  & 65.4  & 71.4  & 63.2  & 58.3  & 64.6  & 70.7  & 73.7  & 66.8  & 61.9  & 68.3  & 14.2$^\ast$ \\
          % & QCLNet\cite{qclnet}(TCSVT'23) &       & 67.9  & 72.5  & 64.3  & 63.4  & 67.0  & 72.5  & 74.8    & 68.5  & 68.9  & 71.2  & 2.6 \\
          & ABCNet\cite{abcnet_Huang_2023_CVPR}(CVPR'23) &       & 65.3  & 72.9  & 65.0  & 59.3  & 65.6  & 71.4  & 75.0   & 68.2  & 63.1  & 69.4  & - \\
          & DAM (Ours) &       & \underline{71.3} & 72.4  & \underline{66.9}  & 61.9  & \textbf{68.1}    & 74.9  & 75.5  & \underline{75.3}  & \textbf{69.8}  & \textbf{73.9}  & \textbf{0.68} \\
          % & \rowcolor{lightgray} Our\_standard &       & 70.3  & \textbf{73.7} & 63.2  & \textbf{66.4} & \textbf{68.4} & \textbf{80.2} & 73.2  & 75.9  & 64.7  & \textbf{70.8} & 71.2  & 82.7  & 42.0 \bigstrut[b]\\
\midrule

    \multicolumn{14}{c}{COCO-20$^i$} \bigstrut\\
\midrule
          % & PPNet(ECCV'20)\cite{ppnet} &       & 36.5  & 26.5  & 26.0 & 19.7  & 27.2  & - & 48.9  & 31.4  & 36.0 & 30.6  & 36.7  & - & 31.5 \bigstrut[t]\\
          % & RPMM(ECCV'20)\cite{pmm} &       & 29.5  & 36.8  & 28.9  & 27.0 & 30.6  & - & 33.8  & 42.0 & 33.0 & 33.3  & 35.5  & - & - \\
          % & RePRI(CVPR'21)\cite{repri} &       & 31.2  & 38.1  & 33.3  & 33.0  & 34.0  & 38.5  & 46.2  & 40.0 & 43.6  & 42.1 & - \\
          & CWT(ICCV'21)\cite{cwt} &       & 32.2  & 36.0 & 31.6  & 31.6  & 32.9  & 40.1  & 43.8  & 39.0 & 42.4  & 41.3 & 47.3 \\
          % & CAPL(CVPR'22) & \multicolumn{1}{p{3.71em}|}{Prototype} & - & - & - & - & 39.8  & - & - & - & - & - & 48.3  & - & - \\
          & DCP(IJCAI'22)\cite{dcp} &  & 40.9  & 43.8  & 42.6  & 38.3  & 41.4  & 45.8  & 49.7  & 43.7  & 46.6  & 46.5  & 11.3 \\
          & NTRENet(CVPR'22)\cite{ntrenet} &  \multicolumn{1}{p{3.71em}|}{Prototype} & 36.8  & 42.6  & 39.9  & 37.9  & 39.3  & 38.2  & 44.1  & 40.4  & 38.4  & 40.3  & 18.6 \\
          & SSP(ECCV'22)\cite{ssp} &       & 35.5  & 39.6  & 37.9  & 36.7  & 37.4  & 40.6  & 47.0 & 45.1  & 43.9  & 44.1 & 8.7 \bigstrut[b]\\
          & IPMT(NeurIPS'22)\cite{ipmt_liu2022intermediate} &       & 41.4  & \underline{45.1}  & \underline{45.6}  & 40.0  & 43.0  & 43.5  & 49.7 & 48.7  & 47.9  & 47.5 & - \\
          % & RPMG-FSS\cite{rpmg-fss}(TCSVT'23) &       & 38.3  & 41.4 & 39.6  & 35.9  & 38.8  & -  & -    & -  & -  & -  & 36.98 \bigstrut[b]\\
          & TAAM(Neurocom'22)\cite{MAO2022104} &       & 32.6  & 37.6 & 31.8  & 33.8  & 34.0  & 36.7  & 44.9    & 37.3  & 35.7  & 38.6  & - \\
\cmidrule(r){2-14} 
\multicolumn{1}{p{5em}|}{ResNet-50}   & CyCTR(NeurIPS'21)\cite{cyctr} &       & 38.9  & 43.0 & 39.6  & 39.8  & 40.3  & 41.1  & 48.9  & 45.2  & 47.0 & 45.6  & 15.9$^\ast$ \bigstrut[t]\\
          & HSNet(ICCV'21)\cite{hsnet} &       & 36.3  & 43.1  & 38.7  & 38.7  & 39.2  & 43.3  & \underline{51.3}  & 48.2  & 45.0 & 46.9  & 2.6 \\
          & AAFormer(ECCV'22)\cite{aaformer} &       & 39.8  & 44.6  & 40.6  & \underline{41.4}  & 41.6  & 42.9  & 50.1  & 45.5  & \underline{49.2}  & 46.9  & - \\
          & VAT(ECCV'22)\cite{vat} & \multicolumn{1}{p{3.71em}|}{Pixelwise} & 39.0 & 43.8  & 42.6  & 39.7  & 41.3  & 44.1  & 51.1  & 50.2  & 46.1  & 47.9  & 3.2 \\
          & DCAMA(ECCV'22)\cite{dcama} &  & \underline{41.9} & \underline{45.1} & 44.4  & \textbf{41.7}  & \underline{43.3}  &  \underline{45.9}  & 50.5  & \underline{50.7}  & 46.0 & 48.3  & 14.2$^\ast$ \\
          % & CMNet(TMM'22) &       & 48.7  & 33.3  & 26.8  & 31.2  & 35    & - & 49.5  & 35.6  & 31.8  & 33.1  & 37.5  & - &  \\
          % & QCLNet\cite{qclnet}(TCSVT'23) &       & 39.8  & \underline{45.7}  & 42.5  & 41.2  & 42.3  & \underline{46.4}  & \textbf{53.0}    & \underline{52.1}  & 48.6  & \underline{50.0}  & 2.6 \\
          & ABCNet\cite{abcnet_Huang_2023_CVPR}(CVPR'23) &       & \textbf{42.3}  & \textbf{46.2}  & \textbf{46.0}  & 42.0  & \textbf{44.1}  & 45.5  & \textbf{51.7}   & \textbf{52.6}  & 46.4  & \underline{49.1}  & - \\
          & DAM (Ours) &       & 39.8  & 41.0 & 40.1  & 40.7  & 40.4  & \textbf{50.1} & 51.0  & 50.4  & \textbf{49.6}  & \textbf{50.3}  & \textbf{0.68} \\
          % & \rowcolor{lightgray} Our\_standard &       & 41.1  & 44.6  & \textbf{46.3} & \textbf{43.1} & \textbf{43.8} & 69.3  & 48.2  & \textbf{52.7} & \textbf{52.5} & \textbf{51.0} & \textbf{51.1} & \textbf{73.7} & 42.0 \bigstrut[b]\\

\midrule
          & CWT(ICCV'21)\cite{cwt} &       & 30.3  & 36.6  & 30.5  & 32.2  & 32.4  & 38.5  & 46.7  & 39.4  & 43.2  & 42.0 & 66.3 \bigstrut[t]\\
          % & CAPL(CVPR'22) & \multicolumn{1}{p{3.71em}|}{Prototype} & - & - & - & - & 42.8  & - & - & - & - & - & 50.4  & - & - \\
          & NTRENet(CVPR'22)\cite{ntrenet} & \multicolumn{1}{p{3.71em}|}{Prototype} & 38.3  & 40.4  & 39.5  & 38.1  & 39.1  & 42.3  & 44.4  & 44.2  & 41.7  & 43.2  & 18.6 \\
         & SSP(ECCV'22)\cite{ssp} &       & 39.1  & 45.1  & 42.7  & 41.2  & 42.0  & 47.4  & \underline{54.5}  & 50.4  & \underline{49.6}  & 50.2  & 27.7 \bigstrut[b]\\
    \multicolumn{1}{p{5em}|}{ResNet-101}  & IPMT(NeurIPS'22)\cite{ipmt_liu2022intermediate} &       & 40.5  & \underline{45.7}  & \underline{44.8}  & 39.3  & 42.6  & 45.1  & 50.3  & 49.3  & 46.8  & 47.9  & - \\
\cmidrule(r){2-14}          
& HSNet(ICCV'21)\cite{hsnet} &       & 37.2  & 44.1  & 42.4  & \underline{41.3}  & 41.2  & 45.9  & 53.0 & 51.8  & 47.1  & 49.5  & 2.6 \bigstrut[t]\\
          & DCAMA(ECCV'22)\cite{dcama} & \multicolumn{1}{p{3.71em}|}{Pixelwise} & \underline{41.5}  & \textbf{46.2}  & \textbf{45.2}  & \underline{41.3}  & \underline{43.5}  & \underline{48.0} & \textbf{58.0} & \textbf{54.3} & 47.1  & \underline{51.9}  & 14.2$^\ast$ \\
          % & QCLNet\cite{qclnet}(TCSVT'23) &       & 40.0  & 45.5  & \underline{45.1}  & \textbf{43.6}  & \underline{43.6}  & 46.9  & \underline{55.8}  & \underline{53.6}   & \underline{51.1}  & \underline{51.9}  & 2.6 \\
          & DAM (Ours) &       & \textbf{44.7} & 44.3  & 44.0 & \textbf{41.8}  & \textbf{43.7}  & \textbf{52.6} & 53.3  & \underline{53.5}  & \textbf{52.8} & \textbf{53.1} & \textbf{0.68} \\
          % & \rowcolor{lightgray} Our\_standard &       & 44.1  & \textbf{47.7} & \textbf{47.5} & \textbf{43.6} & \textbf{45.7} & \textbf{70.1} & 49.7  & 53.8  & 51.6  & 52.0 & 51.8  & 73.2  & 42.0 \bigstrut[b]\\
    % \hline
    \bottomrule
    \end{tabular}%
  \label{tab:SoT}}}%
\end{table*}%

\begin{figure}[t]
  \centering
  \setlength{\abovecaptionskip}{0.4cm}
  \begin{overpic}[width=1.0\linewidth]{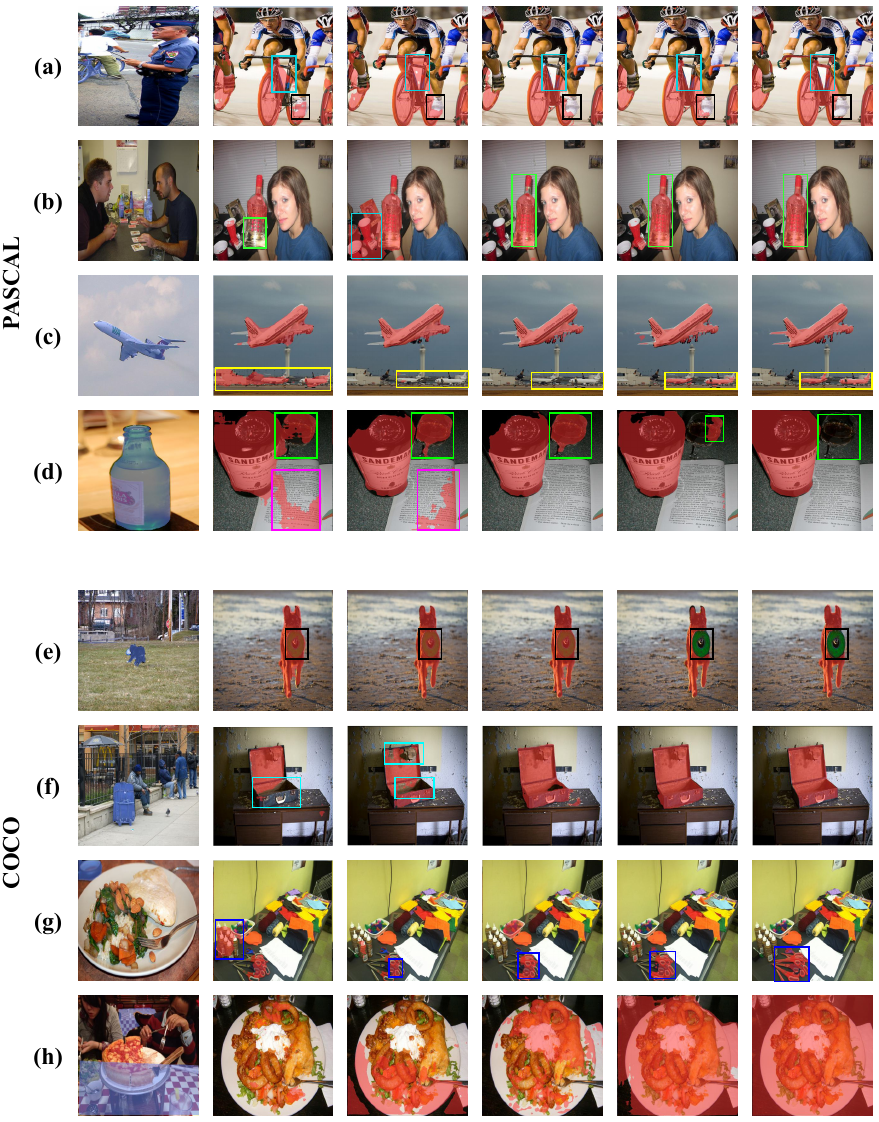}
  \small
  \put(7,-2){{\footnotesize Support}}
   \put(20,-2){{\footnotesize HSNet}}
   \put(30,-2){{\footnotesize DCAMA}}
   \put(42,-2){{\footnotesize DAM$_{\rm{res}50}$}}
   \put(55,-2){{\footnotesize DAM$_{\rm{res}101}$}}
   \put(70,-2){{\footnotesize GT}}
  \end{overpic}
   \caption{The visualization of prediction results with different methods in COCO and PASCAL datasets under various challenging scenarios. `GT' denotes the query ground-truth mask. `DAM$_{\rm{res}50}$' and `DAM$_{\rm{res}101}$' denote DAM with Resnet50 and Resnet101 backbone, respectively. }
   \label{fig:visualize}
\end{figure}

\begin{figure}[t]
  \centering
  \setlength{\abovecaptionskip}{0.4cm}
  \begin{overpic}[width=1.0\linewidth]{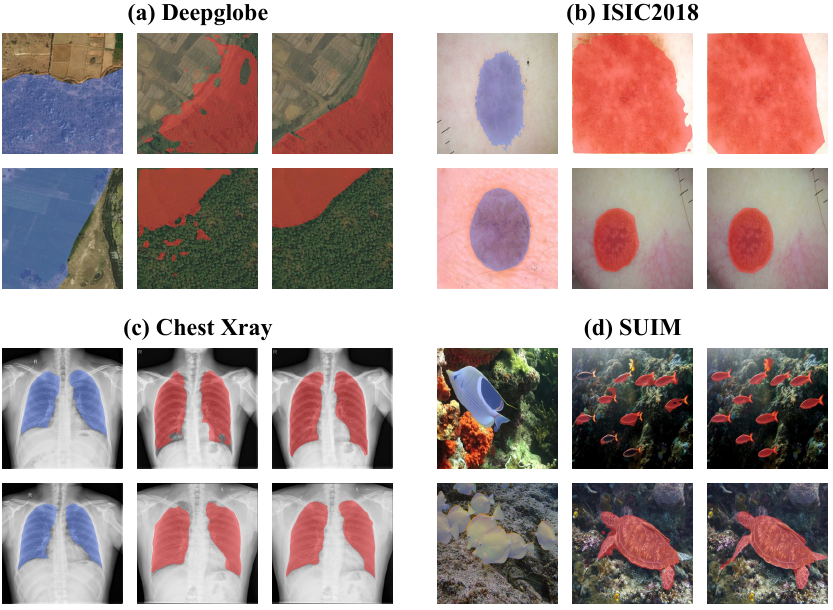}
  \small
  \put(0,-2){Support}
   \put(16,-2){Predicted}
   \put(37,-2){GT}
   \put(52,-2){Support}
   \put(68,-2){Predicted}
   \put(90,-2){GT}
  \end{overpic}
   \caption{The visualization of prediction results in 4 cross-domain tasks. `GT' denotes the query ground truth. }
   \label{fig:cross-visual}
\end{figure}

% including (a) multiple target objects, (b) complicated background, (c) small target object, (d) object with interference, (e) object is covered, (f) full segmentation, (g) object details, and (h) huge object

\begin{table*}[t]
  \centering
  \caption{FSS performances (\%) on cross-dataset task, COCO-$20^i$ $\rightarrow$ Pascal-$5^i$, with different backbones (ResNet50 and ResNet101). The results of all the competitors are from the published literature. The best and the second best results are marked in \textbf{bold} and \underline{underline}, respectively.}
  \makebox[\textwidth]{
  \resizebox{\linewidth}{!}{
    \begin{tabular}{c|l|cccc|c|cccc|c|c}
    \hline
    \multicolumn{13}{c}{COCO20$^i$ $\rightarrow$ Pascal5$^i$} \bigstrut\\
    \hline
    \multicolumn{1}{c|}{\multirow{2}[4]{*}{Backbone}} & \multicolumn{1}{c|}{\multirow{2}[4]{*}{Methods}} & \multicolumn{5}{c|}{1-shot}   & \multicolumn{5}{c|}{5-shot}   & \multicolumn{1}{p{4.1em}}{Learnable} \bigstrut\\
\cmidrule(r){3-12}          &       & \multicolumn{1}{p{2.75em}}{Fold-0} & \multicolumn{1}{p{2.75em}}{Fold-1} & \multicolumn{1}{p{2.75em}}{Fold-2} & \multicolumn{1}{p{2.75em}|}{Fold-3} & \multicolumn{1}{p{2em}|}{mIoU} & \multicolumn{1}{p{2.75em}}{Fold-0} & \multicolumn{1}{p{2.75em}}{Fold-1} & \multicolumn{1}{p{2.75em}}{Fold-2} & \multicolumn{1}{p{2.75em}|}{Fold-3} & \multicolumn{1}{p{2em}|}{mIoU} & \multicolumn{1}{p{4.1em}}{Params(M)} \bigstrut\\
    \hline
          % & RPMM(ECCV'20)\cite{pmm} & 36.3  & 55.0  & 52.5  & 54.6  & 49.6  & 40.2  & 58.0 & 55.2  & 61.8  & 53.8  & - \bigstrut[t]\\
          & PFENet\cite{pfenet} & 43.2  & 65.1  & 66.5  & 69.7  & 61.1  & 45.1  & 66.8  & 68.5  & 73.1  & 63.4  & 34.3 \\
          & RePRI\cite{repri} & 52.2  & 64.3  & 64.8  & 71.6  & 63.2  & 56.5  & 68.2  & 70.0  & 76.2  & 67.7  & - \\
          & HSNet\cite{hsnet} & 45.4  & 61.2  & 63.4  & 75.9  & 61.6  & 56.9  & 65.9  & 71.3  & 80.8  & 68.7  & 2.6 \\
    ResNet-50 & VAT\cite{vat} & 52.1  & 64.1  & 67.4  & 74.2  & 64.5  & 58.5  & 68.0  & 72.5  & 79.9  & 69.7  & 3.2 \\
          & HSNet-HM\cite{hm} & 43.4  & \underline{68.2}  & \textbf{69.4}  & \textbf{79.9}  & 65.2  & 50.7  & \underline{71.4}  & \textbf{73.4}  & \textbf{83.1}  & 69.7  & - \\
          & VAT-HM\cite{hm} & 48.3  & 64.9  & 67.5  & \underline{79.8}  & 65.1  & 55.6  & 68.1  & 72.4  & \underline{82.8}  & 69.7  & - \\
          & RTD\cite{rtd_Wang_2022_CVPR} & \underline{57.4}  & 62.2  & \underline{68.0}  & 74.8  & \underline{65.6}  & \underline{65.7}  & 69.2  & 70.8  & 75.0  & \underline{70.1}  & - \\
          % & DCAMA(ECCV'22) & 66    & 71.8  & 72.2  & 67.1  & 69.3  & 69.3  & 76    & 73.5  & 72.6  & 72.9  & 14.2 \\
          & DAM (Ours) & \textbf{68.8}  & \textbf{70.0}    & 65.1  & 62.3  & \textbf{66.6}  & \textbf{73.9}  & \textbf{74.5}  & \underline{73.3}  & 72.1  & \textbf{73.4}  & \textbf{0.68} \\
          % & OUR\_standard & 70    & 74.2  & 67.9  & 68.5  & 70.2  & 72.7  & 75.7  & 75.2  & 72.2  & 74    & 42 \bigstrut[b]\\
    \hline
          & HSNet\cite{hsnet} & 47.0  & 65.2  & 67.1  & \underline{77.1}  & 64.1  & 57.2  & 69.5  & 72.0  & \underline{82.4}  & 70.3  & 2.6 \bigstrut[t]\\
          & HSNet-HM\cite{hm} & 46.7  & \underline{68.6}  & \textbf{71.1}  & \textbf{79.7}  & 66.5  & 53.7  & 70.7  & \underline{75.2}  & \textbf{83.9}  & 70.9  & - \\
    ResNet-101 & RTD\cite{rtd_Wang_2022_CVPR} & \underline{59.4}  & 64.3  & \underline{70.8}  & 72.0  & \underline{66.6}  & \underline{67.2}  & \underline{72.7}  & 72.0  & 78.9  & \underline{72.7}  & - \\
          % & DCAMA(ECCV'22) & 66.8  & 75.9  & 71.3  & 74.5  & 72.1  & 71.3  & 79.1  & 76.8  & 76.3  & 75.9  & 14.2 \\
          & DAM (Ours) & \textbf{71.0}  & \textbf{72.3}  & 66.6  & 63.8  & \textbf{68.4}  & \textbf{75.2}  & \textbf{76.3}  & \textbf{77.0}  & 72.6  & \textbf{75.3}  & \textbf{0.68} \\
          % & OUR\_standard & 73    & 74.6  & 77.4  & 71.8  & 74.2  & 74.9  & 76.6  & 79.2  & 74.4  & 76.3  & 42 \bigstrut[b]\\
    \hline
    \end{tabular}%
  \label{tab:coco-pascal}}}%
\end{table*}%

\begin{table*}[t]
  \centering
  \caption{FSS performances (\%) on four tasks with Resnet50 backbone under 1-shot and 5-shot settings. All models are trained with the whole PASCAL dataset. Following PATNet\cite{patnet}, \textbf{Deepglobe, ISIC2018, and Chest X-ray are cross-domain tasks, and FSS-1000 is a cross-dataset task.} The best results are marked in \textbf{bold}.}
  \makebox[\textwidth]{
  \resizebox{\linewidth}{!}{
    \begin{tabular}{c|l|cc|cc|cc|cc|cc}
    \hline
    \multirow{2}[4]{*}{Backbone} & \multicolumn{1}{c|}{\multirow{2}[4]{*}{Methods}} & \multicolumn{2}{c|}{Deepglobe} & \multicolumn{2}{c|}{ISIC2018} & \multicolumn{2}{c|}{Chest X-ray} & \multicolumn{2}{c|}{FSS-1000} & \multicolumn{2}{c}{average} \bigstrut\\
\cmidrule(r){3-12}          &       & 1-shot & 5-shot & 1-shot & 5-shot & 1-shot & 5-shot & 1-shot & 5-shot & 1-shot & 5-shot \bigstrut\\
    \hline
          % & PGNet(ICCV'19) \cite{pgnet_Zhang_2019_ICCV} & 10.73 & 12.36 & 21.86 & 21.25 & 33.95 & 27.96 & 62.42 & 62.74 & 32.24 & 31.08 \bigstrut[t]\\
          & PANet\cite{panet} & 36.6 & \textbf{45.4} & 25.3 & 34.0 & 57.8 & 69.3 & 69.2 & 71.7 & 47.2 & 55.1 \\
          % & CANet(CVPR'19) \cite{canet_Zhang_2019_CVPR} & 22.32 & 23.07 & 25.16 & 28.22 & 28.35 & 28.62 & 70.67 & 72.03 & 36.63 & 37.99 \\
          & RPMMs\cite{pmm} & 13.0 & 13.5 & 18.0 & 20.0 & 30.1 & 30.8 & 65.1 & 67.1 & 31.6 & 32.9 \\
          & PFENet\cite{pfenet} & 16.9 & 18.0 & 23.5  & 23.8 & 27.2 & 27.6 & 70.9 & 70.5 & 34.6 & 35.0 \\
        ResNet50  & RePRI\cite{repri} & 25.0 & 27.4 & 23.3 & 26.2 & 65.1 & 65.5 & 71.0 & 74.2 & 46.1 & 48.3 \\
          & HSNet\cite{hsnet} & 29.7 & 35.1 & 31.2  & 35.1  & 51.9 & 54.4 & 77.5 & 81.0 & 47.6 & 51.4 \\
          & PATNet\cite{patnet} & \textbf{37.9} & 43.0 & 41.2 & 53.6 & 66.6 & 70.2  & 78.6 & 81.2 & 56.1 & 62.0 \\
          & DAM (Ours) & 37.1  & 41.6  & \textbf{51.2}  & \textbf{54.5}  & \textbf{70.4}  & \textbf{74.0}  & \textbf{84.6}  & \textbf{86.3}  & \textbf{60.8} & \textbf{64.1} \\
          % & standard & 34.7  &       & 42.9  & 50.6  & 66.7  & 67.9  & 85.2  & 86.5  & 57.38 &  \bigstrut[b]\\
    \hline
    \end{tabular}%
  \label{tab:cross-domain}}}%
\end{table*}%

\begin{table}[t]
  \centering
  \caption{FSS performances (\%) on FSS-1000 with different backbones (ResNet50 and ResNet101). The best results are marked in \textbf{bold}.}
  \small
    \begin{tabular}{l@{\hspace{0.05cm}}|l@{\hspace{0.05cm}}|c@{\hspace{0.1cm}}c@{\hspace{0.05cm}}|c@{\hspace{0.05cm}}c@{\hspace{0.05cm}}}
    \hline
    \multicolumn{6}{c}{FSS1000} \bigstrut\\
    \hline
    \multirow{2}[4]{*}{Backbone} & \multirow{2}[4]{*}{Methods} & \multicolumn{2}{c|}{1-shot} & \multicolumn{2}{c}{5-shot} \bigstrut\\
\cmidrule(r){3-6}          &       & mIoU  & FB-IoU & mIoU  & FB-IoU \bigstrut\\
    \hline
    \multirow{5}[2]{*}{ResNet50} & HSNet & 85.5  & -     & 86.5  & - \bigstrut[t]\\
          & DCAMA & \textbf{88.2}  & \textbf{92.5}  & 88.8  & 92.9 \\
          & CMNet & 82.5  & -     & 83.8  & - \\
          & Ours & 87.9  & 92.3  & \textbf{89.2}  & \textbf{93.3} \\
          % & OUR\_standard & 88.9  & 93    & 89.6  & 93.6 \bigstrut[b]\\
    \hline
    \multirow{5}[2]{*}{ResNet101} & DAN & 85.2  & -     & 88.1  & - \bigstrut[t]\\
          & HSNet & 86.5  & -     & 88.5  & - \\
          & DCAMA & 88.3  & 92.4  & 89.1  & 93.1 \\
          & Ours & \textbf{89.5}  & \textbf{93.4}  & \textbf{90.8}  & \textbf{94.4} \\
          % & OUR\_standard & 89.8  & 93.6  & 90.5  & 94.2 \bigstrut[b]\\
    \hline
    \end{tabular}%
  \label{tab:fss-1000}%
\end{table}%

\begin{table}[t]
  \centering
  \caption{FSS performances (\%) on cross-dataset task, COCO-$20^i$$\rightarrow$FSS1000, with Resnet50 backbone under 1-shot setting. The best results are marked in \textbf{bold}.}
  \small
    \begin{tabular}{l|c@{\hspace{0.1cm}}c@{\hspace{0.1cm}}c@{\hspace{0.1cm}}c|c}
    \hline
    \multicolumn{6}{c}{COCO-5$^i$ $\rightarrow$ FSS1000} \bigstrut\\
    \hline
     & Fold-0 & Fold-1 & Fold-2 & Fold-3 & mIoU \\
    % \multicolumn{1}{p{3em}|}{Methods} & \multicolumn{1}{p{2.7em}}{Fold-0} & \multicolumn{1}{p{2.7em}}{Fold-1} & \multicolumn{1}{p{2.7em}}{Fold-2} & \multicolumn{1}{p{2.7em}|}{Fold-3} & \multicolumn{1}{p{1.5em}}{mIoU} \\
    \hline
          ASGNet & 76.2  & 72.2  & 72.7  & 71.6  & 73.2 \\
          HSNet & 79.9  & 80.5  & 81.1  & 82.1  & 80.8 \\
          SCL & 81.6  & 78.3  & 77.5  & 74.4  & 78.0 \\
          RTD & 82.2  & 82.6  & 79.6  & 83.4  & 81.9 \\
          % DCAMA & 82.9  & 82.0  & 82.5  & 82.1  & 82.4 \\
          DAM (Ours) & \textbf{85.3}  & \textbf{83.7}  & \textbf{83.2}  & \textbf{84.9}  & \textbf{84.3} \\
          % OUR$_{standard}$ & 82.3  & 84.1  & 81.2  & 84.1  & 82.9 \bigstrut[b]\\
    \hline
    \end{tabular}%
  \label{tab:coco-fss}%
\end{table}%

\begin{table}[t]
  \centering
  \caption{FSS performances (\%) on cross-domain task, PASCAL-$5^i$$\rightarrow$SUIM, with Resnet50 backbone under 1-shot setting. The best results are marked in \textbf{bold}.}
  \small
    \begin{tabular}{l@{\hspace{0.1cm}}|c@{\hspace{0.1cm}}c@{\hspace{0.1cm}}c@{\hspace{0.1cm}}c|c}
    \hline
    \multicolumn{6}{c}{Pascal-$5^i$ $\rightarrow$ SUIM} \bigstrut\\
    \hline
    & Fold-0 & Fold-1 & Fold-2 & Fold-3 & mIoU \\
     % \multicolumn{1}{p{3em}|}{Methods} & \multicolumn{1}{p{2.7em}}{Fold-0} & \multicolumn{1}{p{2.7em}}{Fold-1} & \multicolumn{1}{p{2.7em}}{Fold-2} & \multicolumn{1}{p{2.7em}|}{Fold-3} & \multicolumn{1}{p{1.5em}}{mIoU} \\
     \hline
     ASGNet & 32.4  & 30.9  & 28.9  & 35.2  & 31.9 \\
     HSNet & 30.7  & 30.0  & 27.3  & 27.0  & 28.8 \\
     SCL & 31.3  & 31.2  & 32.2  & 32.5  & 31.8 \\
     RTD & 35.2  & 33.4  & 34.3  & \textbf{36.0}    & 34.7 \\
     DAM (Ours) & \textbf{37.1}  & \textbf{35.1}  & \textbf{35.2}  & 31.7  & \textbf{34.8} \\
    % OUR\_standard &       &       &       &       &  \bigstrut[b]\\
    \hline
    \end{tabular}%
  \label{tab:suim}%
\end{table}%

\subsection{Datasets.} 
In this work, we evaluate DAM with seven datasets.  

\textbf{PASCAL-$5^i$} \cite{everingham2010pascal} is the expansion of PASCAL VOC 2012, which is divided into 4 subsets following \cite{repri}. Each subset contains 15 base classes for training and 5 novel classes for testing. 

\textbf{COCO-$20^i$} \cite{lin2014microsoft} contains 80 common classes in natural scenery. Same as PASCAL-$5^i$, COCO-$20^i$ is divided into 4 subsets, each containing 60 base classes and 20 novel classes. 

\textbf{FSS-1000} \cite{li2020fss} is a natural image dataset, consisting of 1,000 classes and each class has 10 annotated samples. The official split has been used in our FSS experiments. The results are tested on the testing set, containing 240 classes and 2,400 samples. 

\textbf{Deepglobe} \cite{deepglobe_Demir_2018_CVPR_Workshops} is a satellite image dataset, containing 7 classes: urban land, agriculture, rangeland, forest, water, barren, and unknown. Following PATNet \cite{patnet}, each image is cut into 6 pieces. 

\textbf{ISIC2018} \cite{isic2018, ham_Tschandl_2018} is a dataset on dermatoscopic images, containing 2,596 skin cancer screening samples. In PATNet \cite{patnet}, ISIC2018 is divided into three classes, while the classification basis has not been published. In our work, the whole dataset is seen as one class, which is more challenging. 

\textbf{Chest X-ray} \cite{candemir2013lung, chest_2014Automatic} is an X-ray dataset for Tuberculosis, which includes 566 annotated X-ray images, collected from 58 cases with a manifestation of Tuberculosis and 80 normal cases. 

\textbf{SUIM} \cite{suim_islam2020semantic} is an underwater imagery dataset, containing over 1,500 pixel-annotated images for eight classes. 

\subsection{FSS Tasks.}
On the basis of the different distributions of the training dataset and testing dataset, the FSS tasks are divided into three types, cross-category, cross-dataset, and cross-domain. \textbf{(1) Cross-category setting} considers a scenario where both the base categories and test categories are sampled from the same dataset, i.e., Pascal-$5^i$, COCO-$20^i$, and FSS1000. \textbf{(2) Cross-dataset setting} evaluates the model trained with one dataset on the other dataset without fine-tuning, i.e., COCO-$20^i\rightarrow$FSS1000, COCO-$20^i\rightarrow$Pascal-$5^i$, and Pascal$\rightarrow$FSS1000. Note that the trained dataset and the evaluated dataset follow either the same distribution or different distributions. In this work, the cross-dataset setting refers to both the trained and evaluated datasets following the same distribution if not specified. \textbf{(3) Cross-domain setting} is a specific case of cross-dataset setting where the trained and evaluated datasets are from different domains, i.e., Pascal$\rightarrow$ISIC2018, Pascal$\rightarrow$Chest Xray, Pascal$\rightarrow$Deepglobe, and Pascal$\rightarrow$SUIM. This is a more challenging setting as the model not only deals with the novel classes but also has to address the domain gap among different datasets. 

% Following the popular FSS competitors, we adopt mIOU to evaluate the model. 

% which training and testing on different datasets, break the distribution of intra-dataset, including COCO20$^i\rightarrow$PASCAL$5^i$ and COCO20$^i\rightarrow$FSS1000. However, the testing dataset, collected from the same scenery as the training dataset, makes FSS easier. \textbf{3) Cross-domain tasks} are the most challenging tasks in FSS, focusing the transfer between different domain, such as training on PASCAL and testing in Deepglobe, ISIC2018, Chest X-ray, and SUIM. Our method is tested on all mentioned benchmarks and achieves great performance.

\subsection{Implementation Details.} 
Our model consists of two parts, backbone and segmentation head. Following the previous works, we conduct experiments on two backbones, i.e., ResNet50 and ResNet101 \cite{he2016deep}, which are pre-trained on ImageNet and frozen during the training stage. The segmentation head is trained with the SGD optimizer in a 0.001 learning rate and 0.9 momentum. We train 100 epochs for PASCAL and FSS datasets, and 50 epochs for the COCO dataset. The batch size is set to 24 for both datasets. The images from both datasets are resized to 384 $\times$ 384. Our model is trained on the PyTorch with two NVIDIA Tesla A100 GPUs.

\subsection{Comparison with State-of-the-Art}
% To evaluate the effectiveness of our method, we compare our method with the state-of-the-art (SOTA) competitors under two few-shot settings in cross-category, cross-dataset, and cross-domain tasks. 

% In this section, we compare our approach with the recent competitors under different settings . 

\textbf{Cross-category Task.} \cref{tab:SoT} shows the comparison results on both PASCAL-$5^i$ and COCO-$20^i$ with two different backbones. With Resnet50 backbone, DAM achieves 71.3\% in PASCAL-$5^i$ and 50.3\% in COCO-$20^i$ under the 5-shot setting, and outperforms the second-best competitors with 1.2\% and 2.0\%, respectively. With Resnet101 backbone, DAM achieves the best performance under both 1-shot and 5-shot settings in PASCAL-$5^i$ and COCO-$20^i$ datasets, outperforming the second-best competitors with 0.2\% and 0.8\% in PASCAL$5^i$, 0.2\% and 1.2\% in COCO$20^i$, respectively. Besides, we observe that DAM has the fewest parameters.

\begin{figure*}[t]
  \centering
  \includegraphics[width=1\linewidth]{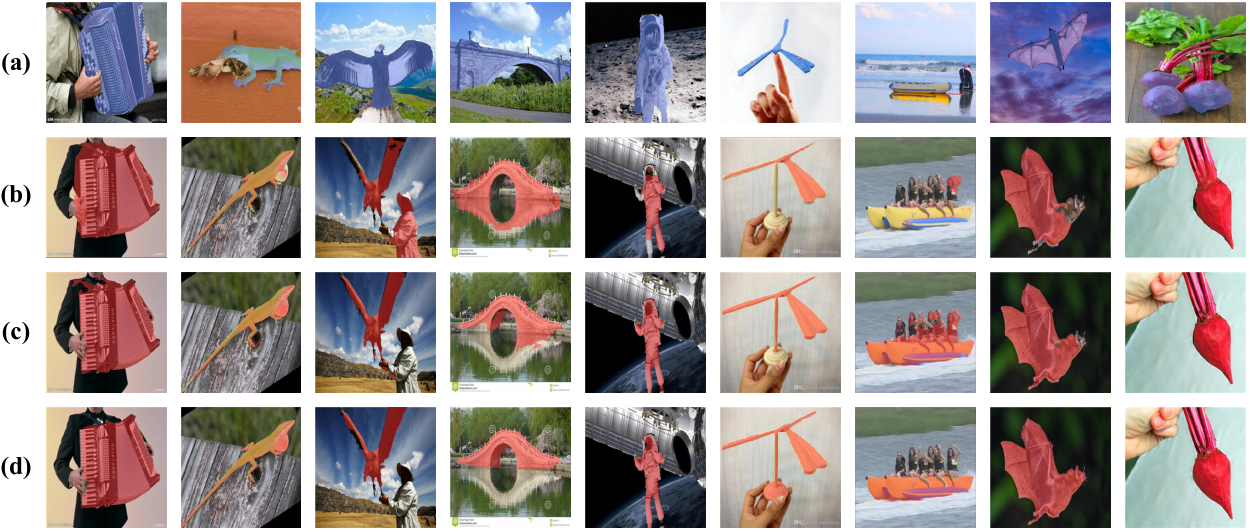}
  \caption{Some 1-Shot visualization on the FSS-1000 dataset with ResNet50 backbone. (a) Support image, (b) HSNet \cite{hsnet}, (c) DAM (Ours), (d) Ground-Truth.}
  \label{fig:fss_compare}
\end{figure*}

As shown in \cref{tab:fss-1000}, our method obtains the best performance with Resnet101 backbone on the FSS-1000 dataset, achieving 1.2\% and 1.7\% improvements under 1-shot and 5-shot settings, respectively. With Resnet50 backbone, our method achieves 87.9\% under the 1-shot setting, as the second-best competitor, while the 5-shot result achieves 89.2\%, surpassing the best competitor DCAMA by 0.4\%. As shown in \cref{fig:fss_compare}, HSNet is underfitting in many samples while our method performs better in all shown samples.

% Compared with all methods mentioned in \ref{tab:SoT}, our method has the fewest parameters. 

% Due to the lightweight design, our method performs weakness in 1-shot setting with Resnet50 backbone, but still achieves 65.4\% in PASCAL$5^i$ and 40.4\% in COCO$5^i$. 

To visualize the segmentation results, some qualitative results of the 1-shot segmentation task are provided in \cref{fig:visualize}. From the results, we observe that our method with Resnet50 segments well in (a) multiple targets, (b) complicated background, (f) full segmentation, and (g) object details, but performs not so well in (c) small objects, (d) objects with interference, (e) object is covered, and (h) huge object. These challenging cases cause performance to decline. In contrast, our method with Resnet101 segments well under all challenging scenarios. Owing to B3DConvs connecting the support and query object regions with feasible sizes, DAM could segment query objects with different sizes to support objects.

\textbf{Cross-dataset tasks.} We conduct experiments on 3 cross-dataset tasks. As shown in \cref{tab:coco-pascal}, DAM achieves the best performance in all settings of COCO-$20^i$ $\rightarrow$ PASCAL-$5^i$ task, surpassing the second-best competitors with 1.0\% and 3.3\% under 1-shot and 5-shot settings with Resnet50 backbone, 1.8\% and 2.6\% under 1-shot and 5-shot setting with Resnet101 backbone, respectively. For the COCO-$20^i$ $\rightarrow$ FSS1000 task in \cref{tab:coco-fss}, DAM achieves 84.3\% as the best performance and surpasses the second-best competitors by 2.4\%. Following PATNet \cite{patnet}, the results of the PASCAL $\rightarrow$ FSS1000 task are shown in the sixth column of \cref{tab:cross-domain}. With Resnet50 backbone, DAM achieves 84.6\% and 86.3\%, outperforming the second-best competitors by 6.0\% and 5.0\% in 1-shot and 5-shot settings, respectively. The significant advantages show the effectiveness of DAM for the cross-dataset FSS tasks.

\textbf{Cross-domain tasks.} We conduct experiments on cross-domain tasks with 4 datasets, which are collected from different scenarios, not intersecting with the trained dataset PASCAL. As shown in \cref{tab:cross-domain}, DAM achieves the best performance under 1-shot and 5-shot settings on both ISIC2018 and Chest Xray datasets, surpassing the second-best competitors by 10.0\% and 0.9\% in ISIC2018, 3.8\% and 4.8\% in Chest Xray, respectively. On the Deepglobe dataset, DAM achieves the second-best performance with 37.1\% and 41.6\% under 1-shot and 5-shot settings, respectively. Following RTD \cite{rtd_Wang_2022_CVPR}, the results of the PASCAL-$5^i$$\rightarrow$SUIM task are shown in \cref{tab:suim}. Our DAM performs slightly better than the second-best competitor with Resnet50 backbone under the 1-shot setting. The results on both \cref{tab:cross-domain} and \cref{tab:suim} demonstrate the effectiveness of the proposed approach on cross-domain FSS tasks. As shown in \cref{fig:cross-visual}, the segmentation results of four cross-domain tasks show that our method performs well in satellite, dermatoscopic, X-ray, and underwater scenarios.

% Some segmentation results of the cross-domain tasks under the 1-shot setting with Resnet50 backbone are visualized in \ref{fig:cross-visual}. 

\begin{table}[t]
  \centering
  \caption{Impact (\%) of B3DConvs and the Support Background under 1-shot setting. `B3D' and `BG' denote the B3DConvs module and the Support Background, respectively. `w 4DConv' denotes exchanging the B3DConvs with 4DConv in HSNet\cite{hsnet}. } %`w/o' denotes removing the part in DAM.
  \small
    \begin{tabular}{l|c@{\hspace{0.1cm}}c@{\hspace{0.1cm}}c@{\hspace{0.1cm}}c|c@{\hspace{0.1cm}}|c}
    \hline
     & Fold-0 & Fold-1 & Fold-2 & Fold-3 & Mean & Params \\
    \hline
    w/o B3D  & 67.1  & 71.2  & 58.9  & 58.8  & 64.0 & 0.65 M \\
    % 2d conv & 67.1  & 71.3  & 63.4  & 58.8  & 65.1 \\
    w/o BG  & 59.0 & 65.0 & 54.3  & 51.2  & 57.4 & 0.68 M \\
    \hline
    w   4D  & 67.1 & 71.8 & 60.8  & 59.2  & 64.7 & 0.68 M \\
    w   B3D & 67.3  & 72.0  & 62.4  & 59.9  & 65.4 & 0.68 M \\
    \hline
    \end{tabular}%
  \label{tab:symmetric}%
\end{table}%

\begin{table}[t]
  \centering
  \caption{Impact (\%) of B3D(Convs) in cross-domain tasks under 1-shot setting. } %`w/o' denotes removing the part in DAM.
  \footnotesize
    \begin{tabular}{l|c@{\hspace{0.1cm}}|c@{\hspace{0.1cm}}|c@{\hspace{0.1cm}}|c|c@{\hspace{0.1cm}}|c}
    \hline
      & Deepglobe & ISIC & Xray & FSS1000 & Mean & Params \\
    \hline
    w/o B3D  & 36.7  & 45.5  & 62.9  & 82.1  & 56.8 & 0.65 M \\
    w   B3D  & 37.1  & 51.2  & 70.4  & 84.6  & 60.8 & 0.68 M \\
    \hline
    \end{tabular}%
    \vspace{-4mm}
  \label{tab:crossdomain}%
\end{table}%

\begin{table}[t]
  \centering
  \caption{Impact (\%) of low-level feature representations from different sources under 1-Shot setting with Resnet50 backbone on the PASCAL dataset. `None' denotes that no low-level feature representations are fused into the decoder. `Both' denotes that both the Support and Query low-level feature representations are fused into the decoder. }
  \small
    \begin{tabular}{l|c@{\hspace{0.1cm}}c@{\hspace{0.1cm}}c@{\hspace{0.1cm}}c|c}
    \hline
    Sources & Fold-0 & Fold-1 & Fold-2 & Fold-3 & Mean \bigstrut \\
    \hline
    None & 66.1 & 68.1 & 60.6 & 54.9 & 62.4 \bigstrut[t] \\
    Support & 64.2  & 67.4  & 60.2  & 55.1  & 61.7 \\
    Query & 67.3  & 72.0  & 62.4  & 59.9  & 65.4 \\
    Both & 67.3  & 70.5  & 62.3  & 57.5  & 64.4 \bigstrut[b]\\
    \hline
    \end{tabular}%
  \label{tab:skip-connect}%
\end{table}%

\begin{table}[t]
  \centering
  \caption{Impact (\%) of Conv1 and Conv2 under 1-Shot setting with Resnet50 backbone on the PASCAL dataset. \textbf{`Params' only indicates the parameters of Conv1 and Conv2.} `ori Conv' and `new Conv' indicate the different Conv1 and Conv2 in left and right of \cref{fig:conv}, respectively.} %`w/o' denotes removing the part in DAM.
  \footnotesize
    \begin{tabular}{l|c@{\hspace{0.1cm}}c@{\hspace{0.1cm}}c@{\hspace{0.1cm}}c|c@{\hspace{0.1cm}}|c}
    \hline
     & Fold0 & Fold1 & Fold2 & Fold3 & Mean &  Convs' Params \\
    \hline
    ori Conv  & 67.3 & 71.2  & 60.1  & 59.3  & 64.5 & 0.161 M \\
    new Conv  & 67.3 & 72.0  & 62.4  & 59.9  & 65.4 & 0.080 M \\
    \hline
    \end{tabular}%
    \vspace{-4mm}
  \label{tab:convlayer}%
\end{table}%

\begin{table}[t]
  \centering
  \caption{Impacts (\%) of `Linear Head' under 1-shot setting with Resnet50 backbone in 10 benchmarks. `w' or `w/o' denote our model (DAM) with or without `Linear Head'. }
  \small
    \begin{tabular}{l|c|cc}
    \hline
    Benchmarks & SOTA & w/o  & w  \bigstrut\\
    \hline
    \multicolumn{4}{c}{Cross-Category} \bigstrut\\
    \hline
    PASCAL$5^i$ & 66.8 & 65.4  & 66.2(+0.6) \bigstrut[t]\\
    COCO$5^i$  & 43.3 & 40.4  & 44.2(+3.8) \\
    FSS1000 & 88.2 & 87.9  & 88.7(+0.8) \\
    \hline
    \multicolumn{4}{c}{Cross-Dataset} \bigstrut\\
    \hline
    COCO$5^i$-PASCAL$5^i$ & 65.6 & 66.6  & 71.4(+4.8) \\
    COCO$5^i$-FSS1000 & 81.9 & 84.3  & 82.6(-1.7) \\
    PASCAL-FSS1000 & 78.6 & 84.6  & 84.1(-0.5) \bigstrut[b]\\
    \hline
    \multicolumn{4}{c}{Cross-Domain} \bigstrut\\
    \hline
    PASCAL-Deepglobe & 37.9 & 37.1  & 36.5(-0.6) \bigstrut[t]\\
    PASCAL-ISIC2018 & 41.2 & 51.2  & 45.5(-5.7) \\
    PASCAL-Chest Xray & 66.6 & 70.4  & 58.3(-12.1) \\
    PASCAL$5^i$-SUIM & 34.7 & 34.8  & 34.7(-0.1) \bigstrut[b]\\
    \hline
    Parameters(M) & 2.6 & 0.68 & 11.2 \\
    \hline
    \end{tabular}%
  \label{tab:linear}%
\end{table}%

\subsection{Ablation Study}
\label{sec:ablation}

In this subsection, we design a series of ablation studies to evaluate the effects of different modules. All the results are obtained with the ResNet50 backbone under the 1-shot task on the PASCAL dataset. 

\textbf{B3DConvs.} \cref{tab:symmetric} illustrates the impact of B3DConvs on the mIoU under 1-shot setting. From the experiment, we observe that B3DConvs brings a 1.4\% improvement to mIoU and a 3.5\% improvement in fold-2, which shows the effectiveness of the B3DConvs module, indicating that exploiting the support-query affinity would be beneficial for FSS performance improvement. When we exchange the B3DConvs module with the 4DConv module in HSNet \cite{hsnet}, there is a 0.7\% dropping in mIoU with ResNet50, which illustrates that the B3DConvs module is more suitable for our DAM.

Furthermore, we conducted an ablation study on the B3DConvs module across four cross-dataset and cross-domain tasks in \cref{tab:crossdomain} under a 1-shot setting. Our results demonstrate that the incorporation of B3DConvs results in additional improvements of 0.4\%, 5.7\%, 7.5\%, and 2.5\% in the Deepglobe, ISIC2018, Chest Xray, and FSS1000 tasks, respectively. This underscores the significance of B3DConvs in enhancing the efficacy of our proposed framework and highlights its potential wider applicability in the field of cross-domain FSS tasks.
% We also provide the ablation study about B3DConvs module in 4 cross-dataset and cross-domain tasks under 1-shot setting in \cref{tab:crossdomain}. From the results, we see that B3DConvs earns extra 0.4\%, 5.7\%, 7.5\%, and 2.5\% in Deepglobe, ISIC2018, Chest Xray, and FSS1000 tasks, respectively. 
% More ablation results about B3DConvs on the other datasets and FSS setting are provided in Supplementary Material. 

\textbf{Support Background.} \cref{tab:symmetric} also illustrates the impact of the background affinity matching on the mIoU. `w/o Background' denotes filtering the background feature from support feature maps with the support mask before the affinity calculation. From the results, removing the background before the affinity would lead to a huge dropping, 8.0\%, in mIoU, which indicates that the support background benefits in the  segmentation of the query sample.

\textbf{Impacts of low-level representations from different sources.} The low-level feature representations are fused into the decoder for query segmentation. In this experiment, we conduct experiments to evaluate the impacts of low-level feature representations from different sources on the final performance. As shown in \cref{tab:skip-connect}, we observe that fusing the low-level feature presentations from the support image would hurt the performance while the low-level presentations from the query image are beneficial for the segmentation results of the query image. Due to the disparity between the support features and the query mask, the classifier head encounters limitations in effectively extracting information from the frozen support features with limited training space.  Consequently, this hampers its ability to support query segmentation and may even introduce training biases.  Conversely, previous studies have demonstrated the efficacy of leveraging lower-level image features for supporting image segmentation.

\textbf{Impacts of Conv1 and Conv2.} The results presented in \cref{tab:convlayer} demonstrate the effectiveness of incorporating a lightweight design for Conv1 and Conv2, yielding an additional 0.9\% improvement under the 1-shot setting in PASCAL with Resnet50. Notably, this design modification also achieves a 50\% reduction in parameters for the Conv Layers. The new designs of Conv1 and Conv2 facilitate the fusion of multi-scale features, a crucial aspect in pixel-level segmentation tasks. Furthermore, the increased number of convolutional layers provides a larger learning space for the affinity matrix to counteract the noise introduced by the frozen backbone during training, which ultimately benefits the segmentation of query samples.

\textbf{Linear Head.} Typically, a linear head is added after each block of the backbone to fit the segmentation task. In this experiment, we conduct experiments to evaluate its impacts on the FSS performances. The experimental results in \cref{tab:linear} show that adding the Linear Head would improve the FSS performance in most cross-class and cross-dataset tasks. Especially in COCO$5^i$, Linear Head takes 3.8\% improvement. However, the model without Linear Head achieves better performance in all cross-domain tasks. We speculate that the linear head facilitates the training of both the cross-category and cross-dataset tasks but may cause the overfitting issue for the cross-domain tasks. To this end, we conclude that adding the linear head on each block of the backbone will boost the FSS performances under both cross-category and cross-dataset scenarios but hurt the performance under the cross-domain scenario. Note that adding the linear head would introduce a large number of parameters.

\begin{table}[t]
  \centering
  \caption{Compared with Segment Anything Model (SAM-base) in two medical image segmentation tasks.}
    \begin{tabular}{c|c|c}
    \hline
          & SAM-base   & Ours \bigstrut\\
    \hline
    Chest Xray & 27.8\%  & 70.4\% \bigstrut[t]\\
    ISIC2018 & 36.1\%  & 51.2\% \bigstrut[b]\\
    \hline
    Parameters & 91M   & 0.68M \bigstrut[t]\\
    datasets & 11M   & 17K  \\
    support image & 0   & 1 \bigstrut[b]\\
    \hline
    \end{tabular}%
  \label{tab:sam}%
\end{table}%

\begin{figure}[t]
  \centering
  \includegraphics[width=1\linewidth]{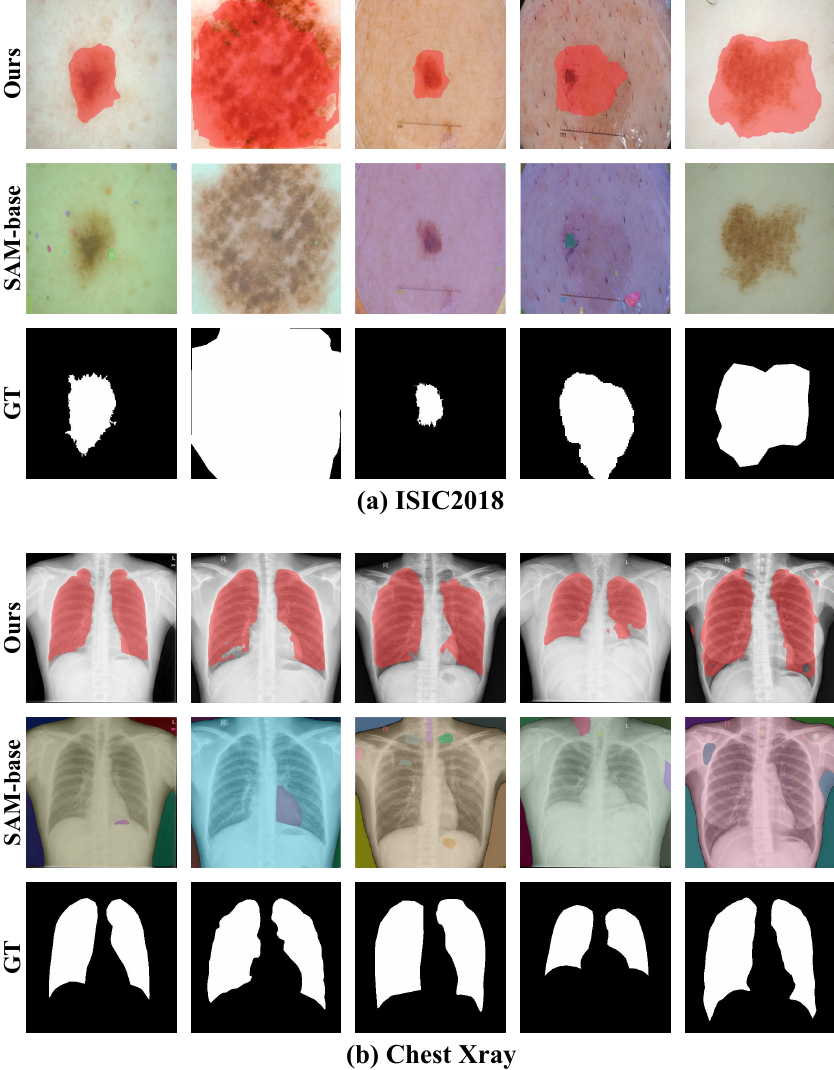}
  \caption{Some visualization of SAM-base on ISIC2018 and Chest X-ray datasets.}
  \label{fig:sam}
\end{figure}

\subsection{Compared With Segment Anything Model (SAM)}
\label{sec:sam}

Recently, there is considerable interest in the implementation of the Segment Anything Model (SAM) \cite{sam_kirillov2023segment}, a large-scale neural network that has garnered attention for its capacity to be trained on datasets consisting of millions of images, billions of annotations, and containing billions of parameters. These works have attained exceptional performance in a multitude of downstream tasks, thereby spurring advancements in the field of computer vision. In this section, we compare SAM and our DAM in the medical segmentation datasets.

% However, our analysis revealed that SAM displays weaker performance in certain medical image segmentation tasks.

In \cref{tab:sam}, we provide the comparison results of our method and SAM-base in two medical image segmentation datasets, i.e., Chest X-ray and ISIC2018. We assess the performance of SAM-base in the "segment everything" setting by calculating the mean Intersection over Union (mIoU) between each mask of segmentation result and target, with the highest achieving score being selected as the measure of segmentation accuracy. The results indicate that our proposed method showcases substantial improvement over SAM-base, despite our method relying on a common Few-Shot Segmentation (FSS) paradigm of segmenting samples with same-category support images. It is worth noting that our model employed in this experiment has fewer parameters and training data than those of the SAM-base. 

Some segmentation results are visualized in \cref{fig:sam}. We observe that SAM-base is unable to effectively segment the lung in the Chest X-ray task, as well as skin cancer in the ISIC2018 task while our DAM segments well on both datasets.

% While in some instances SAM-base was able to detect the presence of skin cancer and segment the surrounding area, it appears that the algorithm lacked an understanding of the skin cancer as the intended target.

% From the experiments, we draw the conclusion that our method, augmented by the use of a support image, demonstrates greater transfer ability in medical image segmentation tasks than SAM-base.

\section{Conclusion}
\label{sec:conclusion}

In this work, we have proposed a lightweight Dense Affinity Matching (DAM) framework for FSS by fully exploiting the relationships between the foreground and background in each support-query pair from both pixel-to-pixel and pixel-to-patch ways, which benefits suppressing the background and highlighting the foreground in query features with a hysteretic spatial filtering module (HSFM). From the results, we conclude that the support background could contribute significantly to the query segmentation by assisting in both filtering the query background and exploring the pixel-to-patch correlations in each support-query pair. Extensive experimental results show that DAM performs very competitively on ten benchmarks under cross-category, cross-dataset, and cross-domain FSS tasks with a small number of parameters.

\section*{Acknowledgment}

This work is supported in part by the National Natural Science Foundation of China under Grant (62002320, U19B2043, 61672456), the Key R\&D  Program of Zhejiang Province, China (2021C01119).

% % Main text
% \section{}\label{}

% % Numbered list
% % Use the style of numbering in square brackets.
% % If nothing is used, default style will be taken.
% %\begin{enumerate}[a)]
% %\item 
% %\item 
% %\item 
% %\end{enumerate}  

% % Unnumbered list
% %\begin{itemize}
% %\item 
% %\item 
% %\item 
% %\end{itemize}  

% % Description list
% %\begin{description}
% %\item[]
% %\item[] 
% %\item[] 
% %\end{description}  

% % Figure
% \begin{figure}[<options>]
% 	\centering
% 		\includegraphics[<options>]{}
% 	  \caption{}\label{fig1}
% \end{figure}

% \begin{table}[<options>]
% \begin{tabular*}{\tblwidth}{@{}LL@{}}
% \toprule
% \midrule
%  & \\
%  & \\
%  & \\
%  & \\
% \bottomrule
% \end{tabular*}
% \end{table}

% % Uncomment and use as the case may be
% %\begin{theorem} 
% %\end{theorem}

% % Uncomment and use as the case may be
% %\begin{lemma} 
% %\end{lemma}

% %% The Appendices part is started with the command \appendix;
% %% appendix sections are then done as normal sections
% %% \appendix

% \section{}\label{}

% % To print the credit authorship contribution details

% \bibliographystyle{cas-model2-names}
\bibliographystyle{elsarticle-num}

\bibliography{egbib}

% % Biography
% % Here goes the biography details.

% % Here goes the biography details.

\end{document}